\documentclass[sigconf]{acmart}

\usepackage{gensymb}
\usepackage{booktabs}
\usepackage{float}

\usepackage{multirow}
\usepackage{longtable}
\usepackage{xspace}
\usepackage{subcaption}

\usepackage{enumitem}

  {\list{}{\selectfont\itshape\leftmargin=0.25in\rightmargin=0.25in}\item[]}%
  {\endlist}

\AtBeginDocument{%
  \providecommand\BibTeX{{%
    \normalfont B\kern-0.5em{\scshape i\kern-0.25em b}\kern-0.8em\TeX}}}

\newcommand{\SystemName}{\textsc{SpaceBlender}\xspace}
\newcommand{\GenericCondition}{\textsc{Generic3D}\xspace}
\newcommand{\TextToRoom}{\textsc{Text2Room}\xspace}
\newcommand{\TextToRoomCondition}{\textsc{Text2Room}\xspace}
\newcommand{\SpaceBlenderCondition}{\textsc{SpaceBlender}\xspace}

\copyrightyear{2024}
\acmYear{2024}
\setcopyright{acmlicensed}\acmConference[UIST '24]{The 37th Annual ACM Symposium on User Interface Software and Technology}{October 13--16, 2024}{Pittsburgh, PA, USA}
\acmBooktitle{The 37th Annual ACM Symposium on User Interface Software and Technology (UIST '24), October 13--16, 2024, Pittsburgh, PA, USA}
\acmDOI{10.1145/3654777.3676361}
\acmISBN{979-8-4007-0628-8/24/10}

\begin{document}

\title{SpaceBlender: Creating Context-Rich Collaborative Spaces Through Generative 3D Scene Blending}

\author{Nels Numan}
\authornotemark[1]
\affiliation{
  \institution{Microsoft Research}
  \country{United States} 
}
\affiliation{
  \institution{University College London}
  \country{United Kingdom} 
}
\email{nels.numan@ucl.ac.uk}

\author{Shwetha Rajaram}
\authornote{This work was done while the first two authors were interns at Microsoft Research. Both authors contributed equally to the paper.}
\affiliation{
  \institution{Microsoft Research}
  \country{United States} 
}
\affiliation{
  \institution{University of Michigan}
  \country{United States} 
}
\email{shwethar@umich.edu}

\author{Balasaravanan Thoravi Kumaravel}
\affiliation{
  \institution{Microsoft Research}
  \country{United States} 
}
\email{bala.kumaravel@microsoft.com}

\author{Nicolai Marquardt}
\affiliation{
  \institution{Microsoft Research}
  \country{United States} 
}
\email{nicmarquardt@microsoft.com}

\author{Andrew D. Wilson}
\affiliation{
  \institution{Microsoft Research}
  \country{United States} 
}
\email{awilson@microsoft.com}

\renewcommand{\shortauthors}{Numan et al.}

\begin{abstract}
There is increased interest in using generative AI to create 3D spaces for Virtual Reality (VR) applications. However, today’s models produce artificial environments, falling short of supporting collaborative tasks that benefit from incorporating the user's physical context. To generate environments that support VR telepresence, we introduce \SystemName, a novel pipeline that utilizes generative AI techniques to blend users' physical surroundings into unified virtual spaces. This pipeline transforms user-provided 2D images into context-rich 3D environments through an iterative process consisting of depth estimation, mesh alignment, and diffusion-based space completion guided by geometric priors and adaptive text prompts. In a preliminary within-subjects study, where 20 participants performed a collaborative VR affinity diagramming task in pairs, we compared \SystemName with a generic virtual environment and a state-of-the-art scene generation framework, evaluating its ability to create virtual spaces suitable for collaboration. Participants appreciated the enhanced familiarity and context provided by \SystemName but also noted complexities in the generative environments that could detract from task focus. Drawing on participant feedback, we propose directions for improving the pipeline and discuss the value and design of blended spaces for different scenarios.
\end{abstract}



\begin{teaserfigure}
  \includegraphics[width=\textwidth]{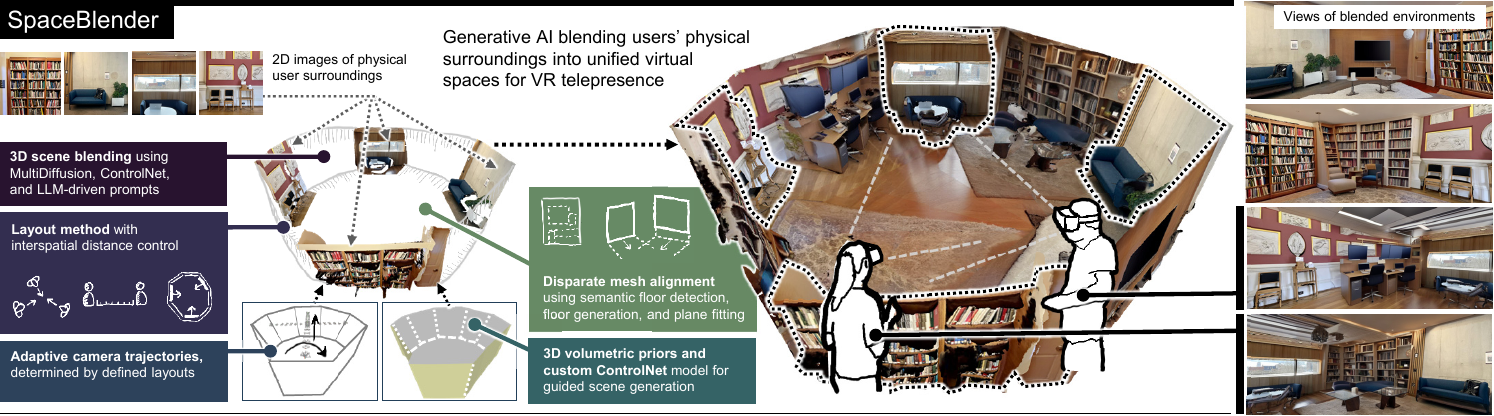}
  \caption{Overview of \SystemName, a pipeline that extends state-of-the-art generative AI models to blend users' physical surroundings into unified virtual environments for VR telepresence.}
  \Description{Description.}
  \label{fig:teaser}
\end{teaserfigure}

\maketitle

\section{Introduction}

There is increased interest in integrating generative models into Virtual Reality (VR) development workflows to accelerate content creation in commercial tools\footnote{\textit{Unity Muse}: https://muse.unity.com, \textit{Bezi AI}: https://www.bezi.com/ai} and enable end-user customization~\cite{delatorreLLMRRealtimePrompting2024,dangWorldSmithIterativeExpressive2023,rajaramBlendScapeEnablingEndUser2024}. 
The recent proliferation of generative AI tools introduces low-effort techniques for end-users to create 3D objects~\cite{poole2022dreamfusion, lin2023magic3d}, panoramic images~\cite{chenText2LightZeroShotTextDriven2022, stanLDM3DLatentDiffusion2023, stanLDM3DVRLatentDiffusion2023, tangMVDiffusionEnablingHolistic2023}, and 3D scenes~\cite{holleinText2RoomExtractingTextured2023, zhouGALA3DTextto3DComplex2024, fangCtrlRoomControllableTextto3D2023, schultControlRoom3DRoomGeneration2024, bautistaGAUDINeuralArchitect2022,wangSceneFormerIndoorScene2021}. Many operate with minimal input such as text and images, offering an easier alternative to the conventional, labor-intensive process of modeling 3D scenes and paving the way for new forms of interactive systems.

In this work, we leverage these developments to explore creating custom virtual environments for VR telepresence systems.
Prior research demonstrated various benefits of incorporating users' familiar real-world context into virtual environments in remote collaboration scenarios, such as supporting deixis~\cite{piumsomboon2018mini, speicher2018360anywhere, orts-escolanoHoloportationVirtual3D2016,gronbaekPartiallyBlendedRealities2023}), mutual awareness \cite{thoravi2022interactive, herskovitzXSpaceAugmentedReality2022, wangSliceLightTransparent2020}, and information recall \cite{das2019memory, krokos2019virtual, essoeEnhancingLearningRetention2022,kochContextReinstatementRequires2024}. Motivated by these findings, we explore a generative approach to creating spaces by blending together multiple users' environmental contexts.
This extends the body of work on aligning dissimilar remote spaces for mixed reality collaboration, e.g., via common object anchors \cite{herskovitzXSpaceAugmentedReality2022, congdonMergingEnvironmentsShared2018, gronbaekPartiallyBlendedRealities2023, kimObjectClusterRegistration2024} or mesh overlays \cite{herskovitzXSpaceAugmentedReality2022, schjerlundOVRlapPerceivingMultiple2022}.

We identify two key challenges with using today's generative AI models to augment the creation of 3D environments for VR telepresence. First, most models are aimed at producing fully synthetic output that is not grounded in real-world spaces \cite{fangCtrlRoomControllableTextto3D2023,fridmanSceneScapeTextDrivenConsistent2023}. Models that attempt to do such grounding require input beyond text and image and can only ground themselves in a single space~\cite{holleinText2RoomExtractingTextured2023, schultControlRoom3DRoomGeneration2024, tangMVDiffusionEnablingHolistic2023}. 
Second, the 3D meshes generated by these models are not explicitly optimized for use as VR environments. 
Through our development process, we found that these generated environments can pose core usability issues for VR telepresence, such as non-navigable pathways, distracting visual and geometric artifacts, and uncanny spaces that detract from user comfort.

To address these challenges, \textbf{we developed \SystemName, a novel pipeline that leverages and extends state-of-the-art generative AI techniques to blend users' physical surroundings into unified virtual spaces suitable for VR telepresence}. This pipeline transforms user-provided 2D images of distinct spaces into context-rich 3D environments through a multi-stage process. First, we transform the 2D images into 3D meshes based on depth estimation, depth alignment, and backprojection. We then employ a RANSAC-based alignment technique to align the disparate 3D meshes, ensuring a uniform floor level. Finally, we use a diffusion-based method for space completion, guided by geometric priors and text prompts defined by a Large Language Model (LLM) acting as an interior architect.

As a preliminary assessment of the suitability of environments generated by \SystemName in supporting collaborative VR, we conducted a comparative study with 20 participants.
In pairs, they performed a VR-based affinity diagramming task in three different virtual environments: \textit{(1)} \GenericCondition: a generic, low-poly room; \textit{(2)} an environment generated with \TextToRoom~\cite{holleinText2RoomExtractingTextured2023}; and \textit{(3)} a \SystemName environment incorporating input images of familiar physical locations provided by participants.
Overall, participants experienced increased physical comfort and navigability in the \GenericCondition and \SpaceBlenderCondition compared to \TextToRoom due to greater consistency in the room geometry. Furthermore, some leveraged recognizable environmental features in the \SpaceBlenderCondition space to complete the clustering task.
While participants envisioned future use cases where incorporating familiar or personal contextual details could provide value, they recommended improvements to the visual quality and realism of \SystemName's environments to better enable these use cases. 

In summary, our work contributes \textit{(1)} the \SystemName generative AI pipeline for creating VR telepresence spaces by blending users' physical surroundings into unified 3D environments; \textit{(2)} a preliminary user study with 20 users that elicited potential benefits, limitations, and use cases of \SystemName, laying the groundwork for future generative AI tools for creating blended environments. 

\section{Related Work}
\label{sec:rw}
\SystemName builds on computational techniques for generating 3D artifacts and prior work that motivates the representation of physical spaces in VR telepresence.

\subsection{Computational Generation of 3D Spaces}
\label{sec:rw:3d-space-generation}
The domain of computational generation of 3D scenes has been significantly revitalized by recent advancements in generative methods. Unlike traditional procedural generation techniques that depend on predefined rules and asset libraries to assemble 3D environments ~\cite{coyneWordsEyeAutomaticTexttoscene2001, vanderlindenProceduralGenerationDungeons2014, sraOasisProcedurallyGenerated2018, chengVRoamerGeneratingOnTheFly2019}, recent approaches can generate entirely new spaces via novel generative techniques. These approaches use a variety of scene representations, including well-known explicit formats like 2D panoramic views~\cite{chenText2LightZeroShotTextDriven2022, stanLDM3DLatentDiffusion2023, stanLDM3DVRLatentDiffusion2023} and meshes~\cite{holleinText2RoomExtractingTextured2023, songRoomDreamerTextDriven3D2023, fangCtrlRoomControllableTextto3D2023, schultControlRoom3DRoomGeneration2024}, as well as recent implicit representations such as Neural Radiance Fields (NeRFs)~\cite{zhangText2NeRFTextDriven3D2023, bautistaGAUDINeuralArchitect2022}, 3D Gaussian Splats~\cite{chungLucidDreamerDomainfreeGeneration2023}, and Signed Distance Functions (SDFs)~\cite{juDiffInDSceneDiffusionbasedHighQuality2024}.

The majority of recent scene generation methods are grounded in the 2D image domain due to the wide availability of image-caption datasets and state-of-the-art generation models trained on these~\cite{rombachHighResolutionImageSynthesis2022}. By leveraging depth estimation models~\cite{bhatZoeDepthZeroshotTransfer2023, baeIrondepthIterativeRefinement2022}, 2D images can be transformed into 3D representations by predicting a depth value for each pixel and backprojecting these values into 3D space. Utilizing models that can handle panoramic images, combined with image upscaling techniques to move beyond image generation models' limited output resolution (e.g.,~\cite{bar-talMultiDiffusionFusingDiffusion2023, kangScalingGANsTexttoImage2023}), systems such as Skybox AI\footnote{\url{https://skybox.blockadelabs.com}} and LDM3D~\cite{stanLDM3DLatentDiffusion2023, stanLDM3DVRLatentDiffusion2023} can generate high-quality $360\degree$ skyboxes based on user-provided text prompts. However, we observe that such techniques generate 3D spaces that do not stay spatially consistent during navigation due to their single-view generation approach, making them less suited for VR telepresence.

Other image-based scene generation methods address this limitation by generating multi-view image sequences~\cite{holleinText2RoomExtractingTextured2023,tangMVDiffusionEnablingHolistic2023,fridmanSceneScapeTextDrivenConsistent2023}. One such method is \TextToRoom, which, given a camera trajectory with matching text prompts and an optional starting image, iteratively expands a 3D mesh through text-conditioned inpainting of 2D rendered views of the mesh~\cite{holleinText2RoomExtractingTextured2023}. In each step, this method estimates a depth image for a rendered view, aligns it with known depth values, and backprojects the depth values to integrate them with the mesh. This is repeated for each viewpoint in the camera trajectory. Finally, an adaptive trajectory is defined at runtime based on a set of viewpoints with the highest number of missing pixels to complete the remaining holes in the mesh. While this approach results in high-quality scenes with improved multi-view consistency for scene navigation, the generated scenes still exhibit structural and visual irregularities, as well as contextual repetitions (e.g., multiple bathtubs in a single bathroom), which limit the navigability and realism required for telepresence.

To address geometric consistency and controllability, \textsc{MVDiffusion}~\cite{tangMVDiffusionEnablingHolistic2023} as well as recent parallel work such as \textsc{Ctrl-Room}~\cite{fangCtrlRoomControllableTextto3D2023} and \textsc{ControlRoom3D}~\cite{schultControlRoom3DRoomGeneration2024} take additional inputs to act as spatial constraints, such as untextured 3D meshes~\cite{tangMVDiffusionEnablingHolistic2023,bautistaGAUDINeuralArchitect2022} or semantic scene maps~\cite{schultControlRoom3DRoomGeneration2024, fangCtrlRoomControllableTextto3D2023}. However, in typical camera-based telepresence systems, such priors are commonly unavailable. Furthermore, like \TextToRoom, these systems are unable to process multiple disparate images. \SystemName advances these concepts by not only accepting multi-image inputs from distinct spaces of collaborating users but also by leveraging the contextual insights from a Visual Language Model (VLM)~\cite{liBLIP2BootstrappingLanguageImage2023}, an LLM~\cite{openaiGPT4TechnicalReport2023}, and a semantic segmentation model~\cite{jainOneFormerOneTransformer2023} to autonomously generate a suitable layout, geometric prior, and text prompts for the generation of an environment directly from user-provided image frames.

\subsection{Physical Spaces in VR Telepresence}

A large body of prior work has studied telepresence systems for supporting users in collaborative tasks. A key objective in these systems is supporting mutual awareness, which refers to the shared understanding of \textit{where} other users are and \textit{what} they are doing~\cite{gutwinWorkspaceAwarenessGroupware1996,dourishAwarenessCoordinationShared1992}. Much of this research focused on establishing awareness in unidirectional settings, where a singular physical space is captured and represented to one or more remote users~\cite{odaVirtualReplicasRemote2015,leeImprovingCollaborationAugmented2017,numanExploringUserBehaviour2022}. This approach supports tasks centered on a single environment, such as remote assistance, by providing remote users with a view into a specific physical space without mutual visibility.

Recent work is increasingly focused on achieving bidirectional awareness to enable new interaction concepts in collaborative settings that not only resemble but also further extend face-to-face collaboration metaphors. These systems often integrate physical and virtual elements belonging to local user’s or remote user’s space into a common interaction space, which is referred to as \textit{Extended Collaborative Space} (\textit{xspace}) by Kumaravel and Hartmann~\cite{thoravi2022interactive}. Based on a literature review, 
\citet{herskovitzXSpaceAugmentedReality2022} identified three categories of techniques for creating such shared spaces: (1) object-centric methods, using specific objects to align spaces~\cite{congdonMergingEnvironmentsShared2018,gronbaekPartiallyBlendedRealities2023,kimObjectClusterRegistration2024}; (2) perspective-driven methods, such as portals~\cite{wangSliceLightTransparent2020,kunertPhotoportalsSharedReferences2014} and world-in-miniature views~\cite{piumsomboonMiniMeAdaptiveAvatar2018,thoravikumaravelLokiFacilitatingRemote2019,lindlbauerRemixedRealityManipulating2018}; and (3) mesh-based methods such as mesh overlays ~\cite{schjerlundOVRlapPerceivingMultiple2022, orts-escolanoHoloportationVirtual3D2016, lindlbauerRemixedRealityManipulating2018}. For example, Loki~\cite{thoravikumaravelLokiFacilitatingRemote2019} used a world-in-miniature volumetric representation to provide real-time awareness cues of remote users’ workspace contexts; RealityBlender~\cite{gronbaekPartiallyBlendedRealities2023} used multiple anchor objects (e.g., whiteboards, tables) to establish an object-centric interaction space; and Slice of Light~\cite{wangSliceLightTransparent2020} used a combination of interactive portals and mesh overlays to enable users to peek into and enter multiple distinct virtual environments.

However, few systems provide \textit{xspaces} that coherently and flexibly include the physical contexts of all users simultaneously. \SystemName aims to enhance such collaborative settings where the inclusion of the physical context of all users is beneficial. Unlike prior systems with distinct boundaries, \SystemName leverages state-of-the-art generative AI models to create cohesively and smoothly blended contextual transitions between disparate spaces. While not explicitly studied in the current work, by incorporating familiar spaces in visually faithful ways, we seek to lay the groundwork for exploring the potential positive effect of these spaces in human information and memory recall \cite{das2019memory, maguire2003routes, dalgleish2013method, legge2012building, krokos2019virtual}.

Several works have employed spatial manipulation techniques to alter or combine spaces, such as Remixed Reality~\cite{lindlbauerRemixedRealityManipulating2018}, which allowed users to make various changes in a live 3D reconstruction of their space, and PointShopAR~\cite{wangPointShopARSupportingEnvironmental2023}, a tablet-based AR tool for modifying 3D point clouds of physical spaces. In contrast, \SystemName automates the customization of captured spaces using images of familiar places to create blended virtual environments, eliminating the need for manual intervention. This approach is particularly relevant for VR telepresence applications, where users often lack the time or ability to manually design or adjust virtual environments for each meeting.

\section{\SystemName System}
\label{sec:system}
This section details the \SystemName pipeline, designed to integrate images of the physical context of multiple users into cohesive virtual environments. \SystemName builds upon the \TextToRoom pipeline due to its extensibility through the usage of off-the-shelf 2D image models, transparent iterative generation process, and ability to initialize generation based on single 2D input images, which are commonly available in telepresence scenarios.

This section starts by outlining the system requirements in Sec.~\ref{sec:sys:objectives}, followed by an overview of the proposed \SystemName pipeline in Sec.~\ref{sec:sys:overview}, and then provides a detailed description of its two phases in Secs.~\ref{sec:system:stage1} and \ref{sec:system:stage2}.

\subsection{Requirements}

\label{sec:sys:objectives}
The assumptions, implementation details, and limitations of the \TextToRoom framework present significant barriers to generating unified spaces from disparate input images for VR usage. Below, we outline the requirements of \SystemName, alongside the associated challenges and our approach to addressing them.

{
\setlength{\parindent}{0cm}
\paragraph{\normalfont\textbf{Requirement 1: Enabling multiple disparate spatial inputs from diverse perspectives and locations.}}
\begin{itemize}[label={}, leftmargin=5pt]\setlength\itemsep{.2em}
    \item \textit{Challenge}: \TextToRoom accepts at most one input image and does not register the resulting mesh in a global coordinate space, while \SystemName must be able to process and align multiple images with various viewpoints to support scene blending.
    \item \textit{Approach}: We introduce a floor plane alignment process that identifies the floor of each mesh by backprojecting semantic values into 3D space, which is then used for global alignment (Sec.~\ref{sec:system:floor_plane_alignment}). For the case where no floor is visible, we propose a technique to synthesize floor sections before alignment. The aligned meshes are then arranged with a parameter-based layout technique (Sec.~\ref{sec:system:layout}).
\end{itemize}

\begin{figure}
    \centering
    \includegraphics[width=\columnwidth]{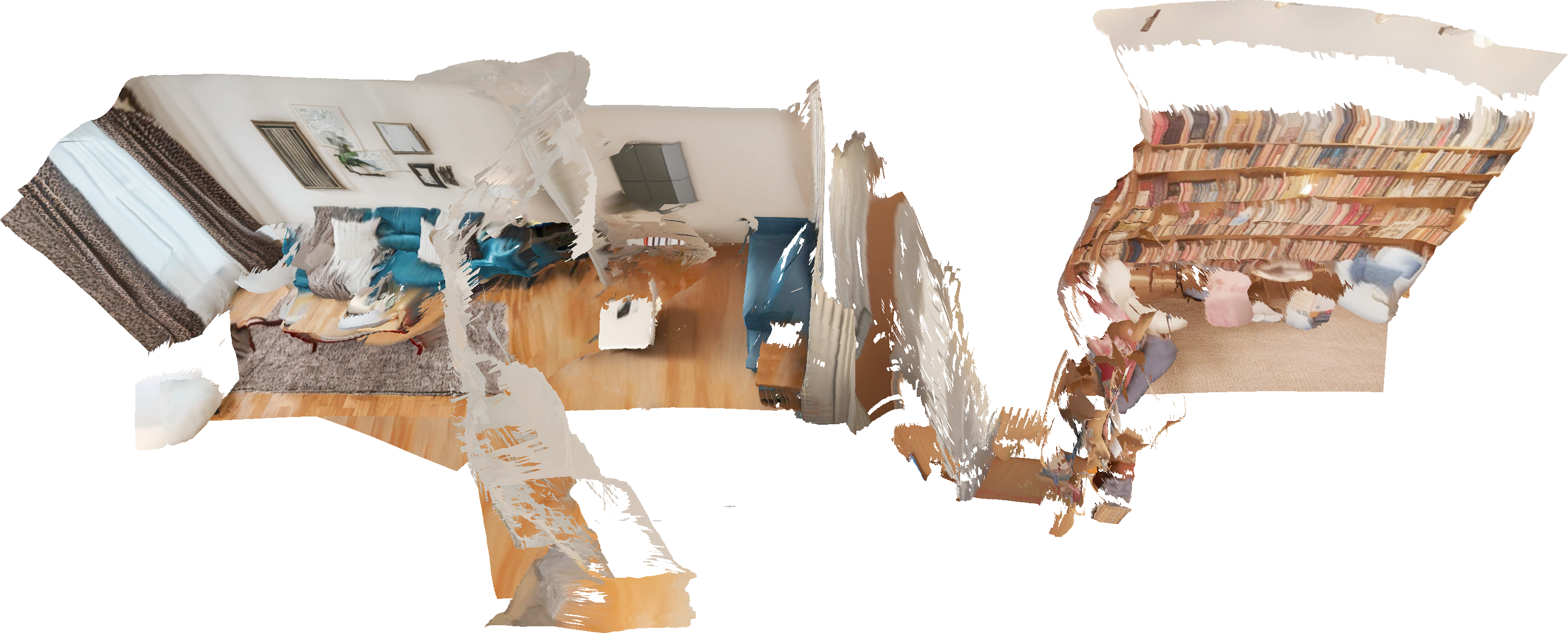}
    \caption{A birds-eye view of two meshes that failed to blend due to the lack of geometric guidance and context throughout the iterative mesh completion process.}
    \label{fig:early-blend-failure}
\end{figure}

\paragraph{\normalfont\textbf{Requirement 2: Enabling coherent scene blending for realistic and context-rich spaces.}}
\begin{itemize}[label={}, leftmargin=5pt]\setlength\itemsep{.2em}
    \item \textit{Challenge}: \TextToRoom uses low-resolution square images in its iterative view inpainting process. This limits the reference frame of the inpainting model, leading to mismatched mesh segments between disparate meshes with harsh geometric and visual boundaries and artifacts, as shown in Fig.~\ref{fig:early-blend-failure}.
    \item \textit{Approach:} We incorporate MultiDiffusion-based~\cite{bar-talMultiDiffusionFusingDiffusion2023} image inpainting to complete wider images, extending the model’s contextual reference window and enabling smooth blends.
    \vspace{.2em}
    \item \textit{Challenge}: \textsc{Text2Room}'s lack of control over room shape causes issues when blending disparate spaces, as shown in Fig.~\ref{fig:early-blend-failure}. 
    \item \textit{Approach:} We propose the usage of a geometric prior defined as the convex hull of the disparate meshes (Sec.~\ref{sec:system:prior}) and a custom ControlNet model for guided scene generation (Sec.~\ref{sec:system:prior-images}).
\end{itemize}

\paragraph{\normalfont\textbf{Requirement 3: Enabling users to create blended environments without the need for extensive manual configuration.}}\setlength\itemsep{.2em}

\begin{itemize}[label={}, leftmargin=5pt]
    \item \textit{Challenge}: The manual configuration required by \TextToRoom for trajectory and prompt adjustments is infeasible for \SystemName's intended application context of VR telepresence for end-users, as the process of trajectory and prompt definition is time-consuming and requires expertise.
    \item \textit{Approach}: We introduce contextually adaptive prompt inference based on a VLM and an LLM (Sec.~\ref{sec:sys:prompt-inference}) as well as adaptive trajectories (Sec.~\ref{sec:sys:completion}), enabling cohesive and automated space blending.
\end{itemize}

\paragraph{\normalfont\textbf{Requirement 4: Supporting core VR usability requirements for end-users including comfortable navigation and viewing.}}\setlength\itemsep{.2em}

\begin{itemize}[label={}, leftmargin=5pt]
    \item \textit{Challenge}: The depth estimator used by \TextToRoom commonly produces slanted and discontinuous floors and walls, which is problematic for VR navigation and spatial orientation. 
    \item \textit{Approach}: To achieve a consistent room structure, \SystemName performs semantic segmentation on inpainted images and copies the depth values for wall, floor, and ceiling pixels from the rendered depth image of the geometric prior. These values are also used to inform depth completion for remaining pixels.
\end{itemize}
}

\begin{figure*}[ht]
    \centering
    \includegraphics[width=\textwidth]{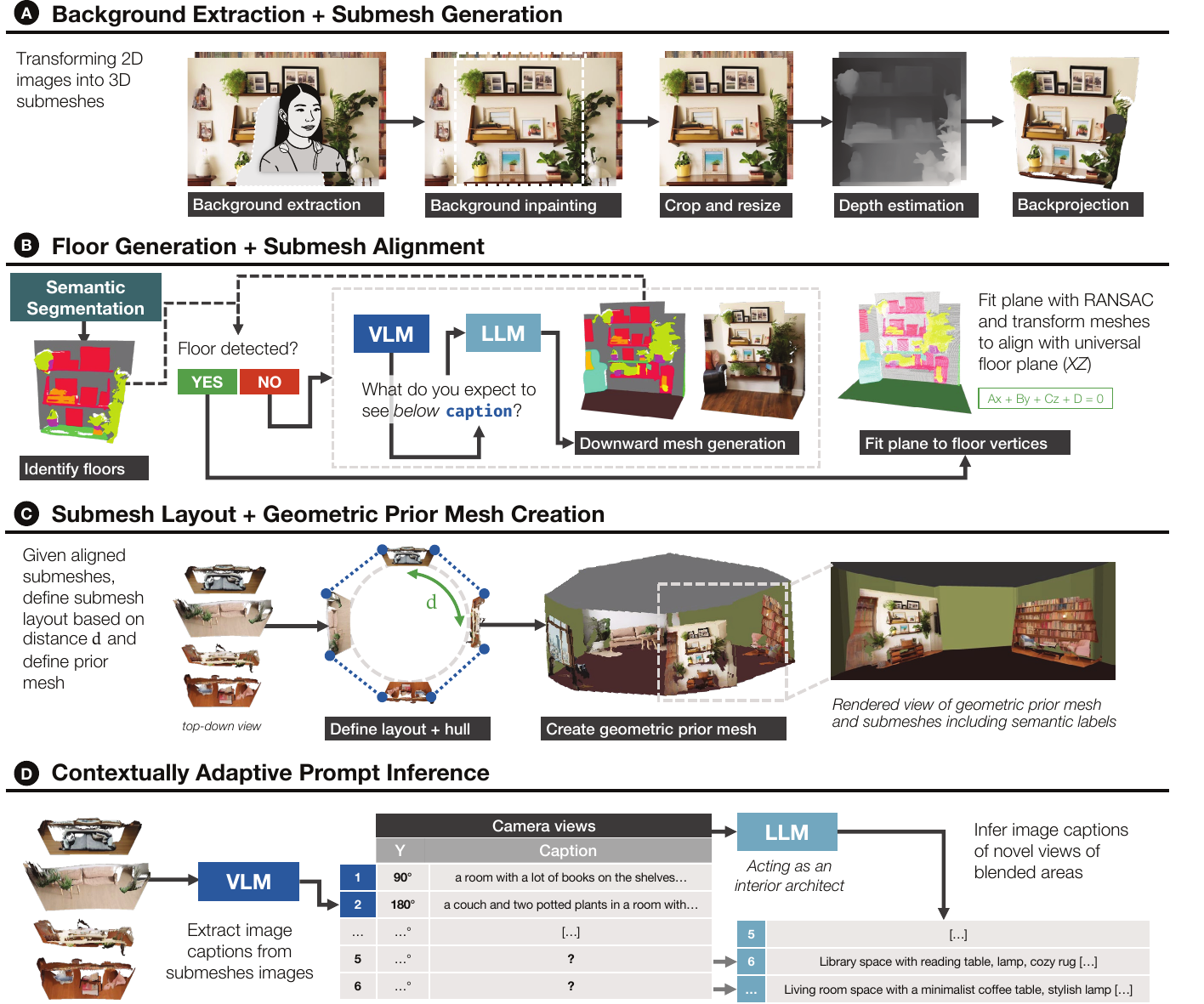}
    \caption{Overview of \textit{Stage 1} components as described in Sec.~\ref{sec:system:stage1}.}
    \label{fig:phase1}
\end{figure*}

\subsection{System Overview}
\label{sec:sys:overview}
The \SystemName pipeline processes $n$ input images to produce a 3D mesh that integrates the context of each image into a cohesive blended environment. The pipeline has two main stages: the first runs once per generation, while the second is iterative, similar to \textsc{Text2Room}. Below, we give a brief overview of these stages, with detailed descriptions available in the following subsections.

In \textit{Stage 1}, each input image is first preprocessed, after which depth values of each pixel are estimated and backprojected to create a 3D mesh (Sec.~\ref{sec:system:2d_to_3d}). We refer to the resulting $n$ meshes as \textit{submeshes} throughout the remainder of this paper. Next, the submeshes are aligned to a common floor plane with a RANSAC-based method applied to floor vertices identified by a semantic segmentation model (Sec.~\ref{sec:system:floor_plane_alignment}), optionally including a floor generation step if no floor is visible in the image. The aligned submeshes are then positioned based on a parameter-based layout technique (Sec.~\ref{sec:system:layout}) based on which a geometric prior mesh is created to define the shape of the blended space (Sec.~\ref{sec:system:prior}). Lastly, text prompts describing the blended regions (i.e., the empty space between submeshes) of the environment are generated with an LLM based on captions inferred by a VLM (Sec.~\ref{sec:sys:prompt-inference}).

In \textit{Stage 2}, the submeshes are blended through a process that involves repeatedly inpainting and integrating 2D rendered views of the mesh. For each iteration, based on the submesh layout defined in \textit{Stage 1}, geometric image priors are rendered to function as a guide for the shape of the space (Sec.~\ref{sec:system:prior-images}). These are combined with the generated text prompts from \textit{Stage 1} to guide the content and appearance of the space (Sec.~\ref{sec:system:blending}). Once the blending process completes, an adaptive mesh completion trajectory is followed to fill remaining gaps in the environment (Sec.~\ref{sec:sys:completion}).

\paragraph{Implementation} Like the original \TextToRoom implementation, we use Stable Diffusion 1.5 \cite{rombachHighResolutionImageSynthesis2022} for image generation and inpainting and IronDepth~\cite{baeIrondepthIterativeRefinement2022} for depth estimation and inpainting. Furthermore, \SystemName uses the BLIP-2~\cite{liBLIP2BootstrappingLanguageImage2023} VLM, GPT-4 \cite{openaiGPT4TechnicalReport2023} LLM, and OneFormer~\cite{jainOneFormerOneTransformer2023} semantic segmentation model. We decoupled the image inpainting process through the usage of a local API A1111 WebUI server API endpoint\footnote{\url{https://github.com/AUTOMATIC1111/stable-diffusion-webui}} to provide enough GPU memory for the usage of the various models in our pipeline. The server and pipeline run on separate machines, each equipped with an NVIDIA RTX 4090 GPU in our local setup. It takes about 55-60 minutes to generate a \SystemName environment with this configuration.

\subsection{\textit{Stage 1}: From 2D Images to 3D Layout}
\label{sec:system:stage1}
The pipeline’s first stage establishes the spatial structure of the blended environment. It begins with image preprocessing and depth estimation, converting 2D images into 3D submeshes. These submeshes are aligned to a common floor plane and arranged in a circle, based on which a geometric prior is defined. Finally, prompts for the blended regions are generated by an LLM.

\subsubsection{From 2D Images to 3D Submeshes (\textbf{Fig.~\ref{fig:phase1}A})}
\label{sec:system:2d_to_3d}

In this step, the $n$ input images are projected into 3D space. First, a semantic segmentation model is used to detect people in each input image. If a person is detected, that region is removed and inpainted using a prompt inferred by the VLM. The resulting image is then cropped to $512 \times 512$ pixels to ensure compatibility with subsequent models in the pipeline. Next, using \textsc{Text2Room}’s existing functionality, a 3D mesh is created. This involves using the depth estimation model to estimate depth values, aligning them with known depth values, and backprojecting them into 3D space along with the colors of the input image.

\subsubsection{Submesh Alignment (\textbf{Fig.~\ref{fig:phase1}B})}
\label{sec:system:floor_plane_alignment}

\begin{figure}
    \centering
    \includegraphics[width=\columnwidth]{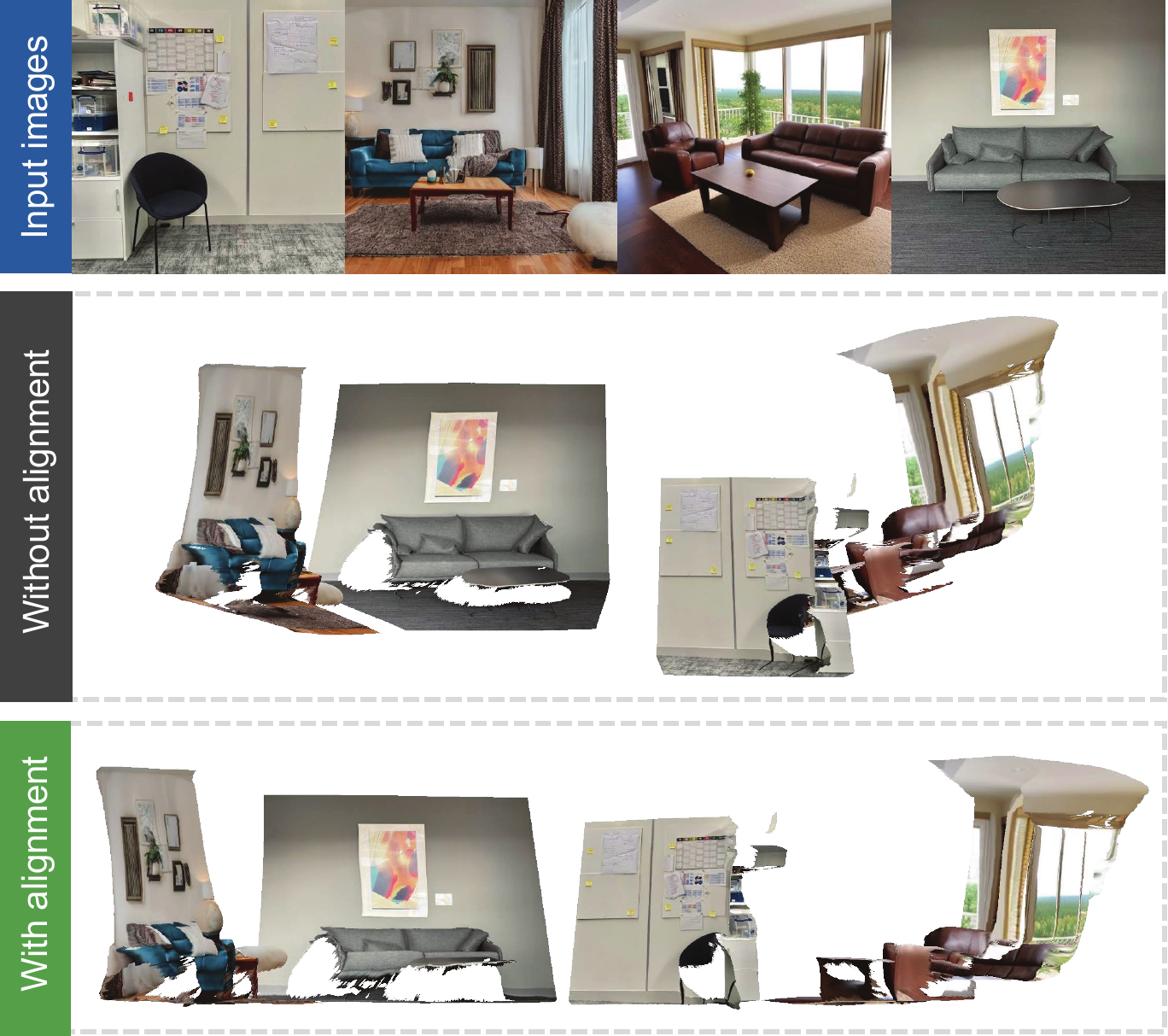}
    \caption{Comparison between unaligned submeshes and submeshes aligned with our semantic floor alignment technique. The unaligned spaces have floors at different levels and inclines that can be jarring to navigate.}
    \label{fig:floor_alignment_comparison}
\end{figure}

We introduce a floor plane alignment technique to reconcile differing perspectives in input images, ensuring the spatial consistency that is needed for scene blending (Fig.~\ref{fig:phase1}B). First, labels for floor-like objects (e.g., floor, carpet) within each submesh are obtained using a semantic segmentation map derived from the input image. These semantic labels are then backprojected into 3D space, replacing the submesh’s RGB colors with semantic label values. To handle any discrepancies between the depth estimation and semantic segmentation model output, floor vertices more than 0.3 meters above or below the median Y-coordinate of the floor-like vertices are excluded.

Next, RANSAC is used to identify a plane corresponding to these floor-like vertices.
To ensure a hypothetical plane is a floor, we use three additional heuristics: (1) the plane's orientation must be within $45\degree$ of the target plane normal; (2) the normal vector must have a positive Y-component; and (3) the extent of the inlier points in the X and Z axes should be at least 0.5 meters.
After selecting the best floor plane candidate, we rotate the mesh to align the plane’s normal with the Y-axis. Next, we translate the floor to $Y=0$ and set the minimum Z-coordinate to 0. Figure~\ref{fig:floor_alignment_comparison} shows an example of submeshes aligned to a common floor plane.

\begin{figure*}
    \centering
    \includegraphics[width=\textwidth]{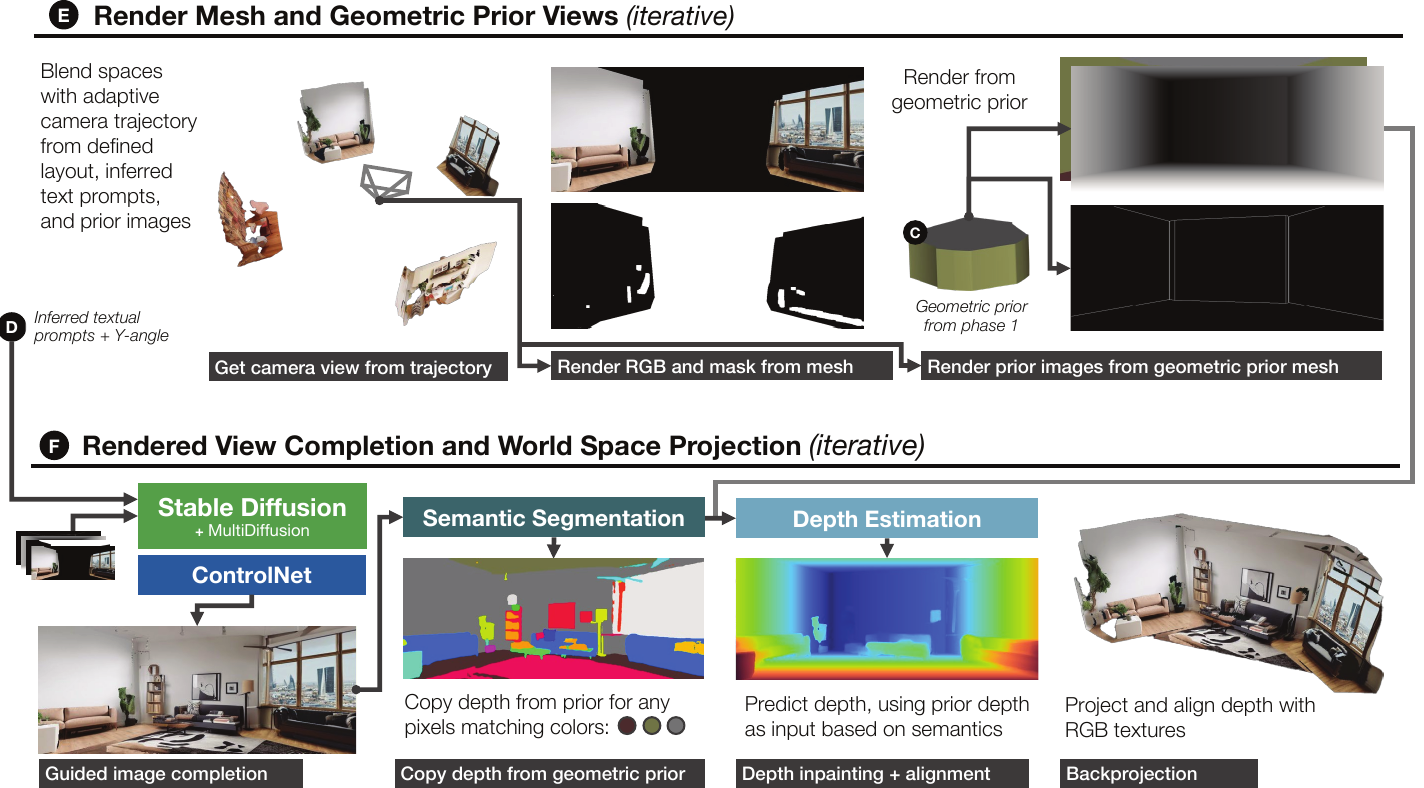}
    \caption{Overview of \textit{Stage 2} components as described in Sec.~\ref{sec:system:stage2}.}
    \label{fig:phase2}
\end{figure*}

\paragraph{Floor Generation (\textbf{Fig.~\ref{fig:phase1}B})}
If input images lack a floor (e.g., only show a wall), preventing floor plane fitting, we employ a generative technique to extend the submesh downward to create a floor for alignment. This method follows a five-step trajectory that interpolates between the following: gradually looking downward (from -5 to -30 degrees), moving backward (from 1 to 1.5 meters), and moving upward (from 0.3 to 1 meter) relative to the initial camera view. For each step, we use a custom floor description generated by the LLM based on the caption of the image obtained with a VLM. If floor generation fails after ten attempts, the submesh remains unaligned for the rest of the generation process.

\subsubsection{Submesh Layout (\textbf{Fig.~\ref{fig:phase1}C})}
\label{sec:system:layout}
With submeshes aligned to a universal floor plane, each submesh is transformed to form a layout resembling an open space. This is achieved by positioning the front side (i.e., the side facing the camera used to capture the input image) of each submesh on the perimeter of a circle as seen from a top-down view. The diameter of this circle is determined by a configurable parameter $d$, which controls the distance between the submeshes. Each submesh faces the center of the unified space, ensuring a clear line of sight between all submeshes. This design choice was made in consideration of the importance of mutual awareness in collaborative scenarios~\cite{gutwinWorkspaceAwarenessGroupware1996,dourishAwarenessCoordinationShared1992}.

\subsubsection{Geometric Prior Mesh (\textbf{Fig.~\ref{fig:phase1}C})}
\label{sec:system:prior}
Given the aligned submeshes, a geometric prior mesh is generated to define the shape of the blended space. This involves the definition of a mesh based on the convex hull of the submesh layout, with faces assigned to represent the floor, walls, and ceiling. The height of this mesh is set to the height of the tallest submesh, or 2.5 meters if none is taller. The floor, ceiling, and wall faces are colored based on their respective semantic label colors from the ADE20K dataset~\cite{zhouSceneParsingADE20K2017}. This mesh is used in rendering prior images for the iterative image inpainting process, as described in Sec.~\ref{sec:system:stage2}.

As the convex hull effectively forms straight walls between submeshes, the number of submeshes and their shapes directly impact the overall shape of the geometric prior, and conversely, the blended space. For example, four submeshes with straight walls create an octagon-like shape, while four submeshes with straight corners result in a square-like shape. A visual explanation and example of spaces with different numbers of input images and submesh shapes are available in Figs.~\ref{fig:prior-influence-explanation} and \ref{fig:prior-influence-examples}, respectively.

\subsubsection{Contextually Adaptive Prompt Inference (\textbf{Fig.~\ref{fig:phase1}D})}
\label{sec:sys:prompt-inference}
In preparation of the iterative mesh generation process in \textit{Stage 2}, text prompts are generated to describe the intended contents of the blended regions. This begins by obtaining image descriptions for each submesh using the VLM, along with a rotation value indicating their relative orientation from a top-down view. This data is then passed to an LLM instructed to act like a highly creative interior architect and photographer skilled at designing spaces with diverse contexts and appearances while avoiding repetitive objects. Based on the rotation values and known image descriptions, the LLM returns a set of new image descriptions paired with rotation values corresponding to the regions to be blended (i.e., the blank regions between submeshes). The full system prompt of this LLM is provided in Sec.~\ref{appendix:system_prompt}.

\subsection{\textit{Stage 2}: Iterative Blending Guided by Geometric Priors and Contextual Prompts}
\label{sec:system:stage2}
Utilizing the submesh layout, geometric prior mesh, and text prompts from Stage 1, this stage iteratively blends the disparate submeshes into a unified environment.

\subsubsection{Room Shape Guidance with Geometric Prior Images (\textbf{Fig.~\ref{fig:phase2}E})}
\label{sec:system:prior-images}

To enable geometrically coherent space blending, \SystemName uses a collection of prior images to guide the iterative text-conditioned image inpainting step of the mesh blending and completion process via ControlNet~\cite{zhangAddingConditionalControl2023}. Each time the process renders a view of the mesh for inpainting, it also renders a set of prior images from the same camera viewpoint based on the geometric prior mesh aligned with the submesh layout. \SystemName is capable of generating three types of prior images which each have a distinct effect on the output of the image inpainting process:

\begin{itemize}
\item \textbf{Depth Prior:} This prior type can act as a hard constraint for generating spaces with shapes and contents similar to the geometric prior. It is defined by rendering the relative depth for a specified view of the geometric prior mesh.
\item \textbf{Layout Prior:} This prior type enables constraining the shape of the space without limiting its content (i.e., furniture). It is defined by calculating depth gradients using the Sobel operator to form surface normals based on the depth prior. Subsequently, the magnitude of these normals is calculated and processed with Canny edge detection to produce an image that effectively outlines the geometric prior mesh with white lines outlining the walls, floor, and ceiling on a black background.
\item \textbf{Semantic Prior:} This prior type can act as an additional hard constraint to guide the semantic contents of the inpainted views. The current geometric prior mesh definition of \SystemName only defines semantic labels for the walls, floor, and ceiling, making it suitable to serve as room layout composition guidance when an empty open space is desired. 
\end{itemize}

\begin{figure*}
    \centering
    \includegraphics[width=\textwidth]{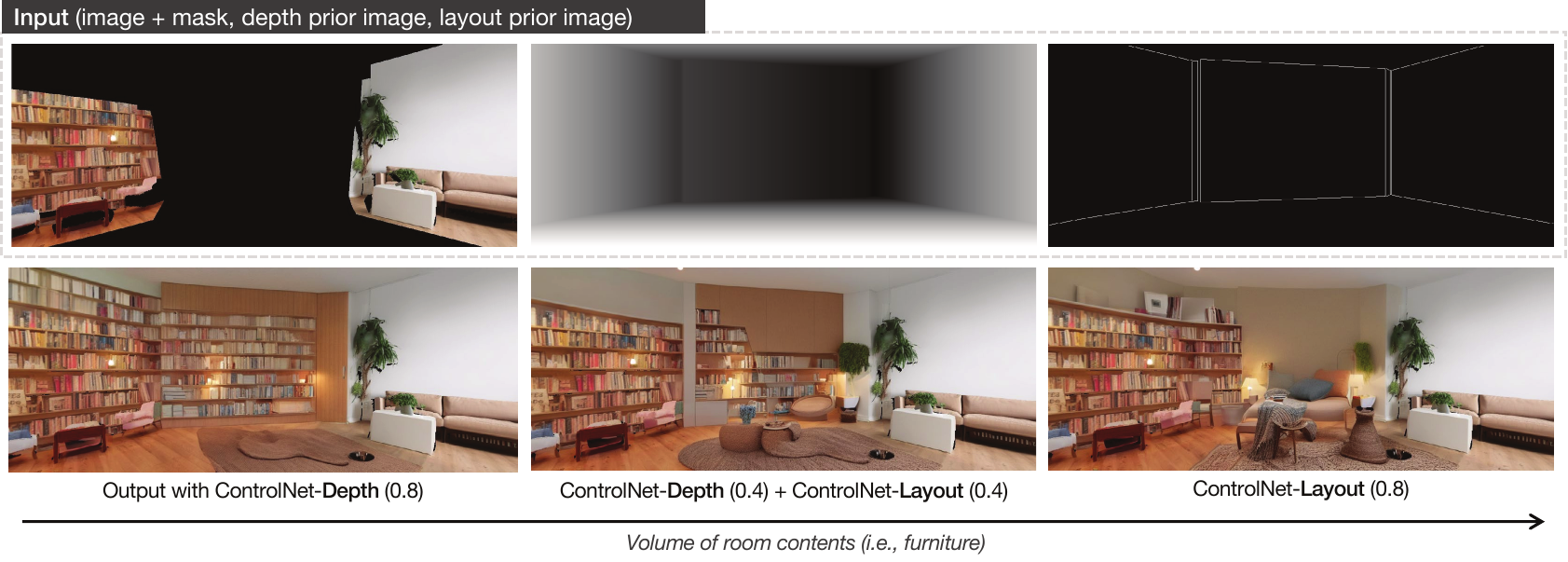}
    \caption{Comparison of output generated with varying weights of ControlNet depth and layout models, impacting the prior's impact on the output (generated with fixed seed). Top: input images including the input image, depth prior image, and layout prior image rendered from geometric prior. Bottom: results with varying weights are indicated in parentheses.}
    \label{fig:layout_prior}
\end{figure*}

These prior images can be stacked and combined using Multi-ControlNet\footnote{\url{https://github.com/Mikubill/sd-webui-controlnet\#multi-controlnet}}, which allows for adjusting each prior’s influence on the image output. For example, using only the layout prior guides the model to generate a space with a specific room structure while allowing the room content (e.g., furniture) to be generated freely. A depth prior can be added to guide the image inpainting model to generate furniture commonly positioned near the walls (e.g., sofas and bookshelves). An example of the depth and layout priors’ influence on the output is shown in Fig.~\ref{fig:layout_prior}, demonstrating the varying effects on room contents, with more examples shown in Fig.~\ref{fig:control_layout_extra_samples}.

While the depth and semantic prior images are used with pre-trained ControlNet models\footnote{\url{https://huggingface.co/lllyasviel/ControlNet}\label{footnote:controlnet-models}}, the layout prior is used with a custom ControlNet model, \texttt{ControlNet-Layout}. We describe the training process of this model below.

\paragraph{Training ControlNet-Layout}
We trained \texttt{ControlNet-Layout} on a dataset of 13,182 images. Instead of using an existing dataset, we created our own by using the pre-trained ControlNet segmentation model\footnotemark[\getrefnumber{footnote:controlnet-models}] using semantic maps inferred from SUN-RGBD~\cite{songSUNRGBDRGBD2015} and LSUN~\cite{zhang2016largescale,yuLSUNConstructionLargescale2016}, resized to $512 \times 512$ pixels. This was repeated several times with unique seeds to enhance image quality and diversify the dataset by generating multiple images per segmentation map. The training process was initialized with the weights of the pre-trained M-LSD model of ControlNet\footnotemark[\getrefnumber{footnote:controlnet-models}] and used a learning rate of $1 \times 10^{-5}$ and a batch size of 4. Training was halted after one epoch due to satisfactory performance on the validation set. Fig.~\ref{fig:controlnet_layout_loosecontrol_comparison} shows examples of \texttt{ControlNet-Layout} output, including a comparison to a similar model from recent parallel work~\cite{bhatLooseControlLiftingControlNet2023}.

\subsubsection{Iterative Space Blending (\textbf{Fig.~\ref{fig:phase2}F})}
\label{sec:system:blending}

This step unifies disparate submeshes iteratively, according to the geometric prior images and prompts defined in Sec.~\ref{sec:sys:prompt-inference}. To enable \SystemName’s blending capabilities, we broaden the context window of the image inpainting model by increasing the resolution from $512 \times 512$ (as used by \TextToRoom) to $512 \times 1280$ while maintaining the rendering camera's field-of-view of $55\degree$. This is enabled by an A1111 WebUI plugin implementation of MultiDiffusion\footnote{\url{https://github.com/pkuliyi2015/multidiffusion-upscaler-for-automatic1111}}~\cite{bar-talMultiDiffusionFusingDiffusion2023}.

By increasing the width of the images generated throughout the blending process, we broaden the inpainting model's environmental reference frame and enable it to blend the space between neighboring submeshes in a single inpainting step (see Fig.~\ref{fig:layout_prior}), yielding higher fidelity results compared to step-wise blending, which results in harsh boundaries and artifacts such as shown in Fig.~\ref{fig:early-blend-failure}.

However, due to its circular layout method, when \SystemName is given three or fewer input images, the large distances between submeshes still prevent blending in a single step. In these cases, \SystemName uses LLM-based prompts to create intermediate submeshes with an image generation model and integrates these with the layout to bridge the gaps between submeshes and enable blending. The rightmost example in Fig.~\ref{fig:prior-influence-explanation} demonstrates this approach.

\subsubsection{Mesh Completion Trajectory}
\label{sec:sys:completion}
The initial iterative blending process creates a mesh that horizontally integrates the submeshes, defining the blended space from a central perspective. However, at this stage, the floor and ceiling are largely incomplete, and the mesh contains significant gaps that need filling. To address this, an additional set of camera trajectories is employed.

First, interpolation-based trajectories are generated to cover the missing sections of the floor and ceiling. Then, trajectories for each submesh are defined. These paths interpolate the position and rotation of the camera viewpoint, starting centrally within the unified space and initially directed at a specific submesh. The interpolation trajectory ends at the submesh center, facing either the left or right adjacent submesh. Throughout this process, the text prompt passed to the image inpainting model is defined as the text prompt of the blended area that is most closely aligned with the camera's view.

Lastly, an additional trajectory simulates a user looking around the unified space from the center point of their submesh, ensuring the mesh accounts for gaps visible from typical user perspectives. To mimic natural gaze variations, a degree of randomness is introduced into these viewpoints. Once all trajectories are completed, the blended space is ready for use in a VR telepresence system.

\begin{figure*}[ht]
    \centering
    \includegraphics[width=\textwidth]{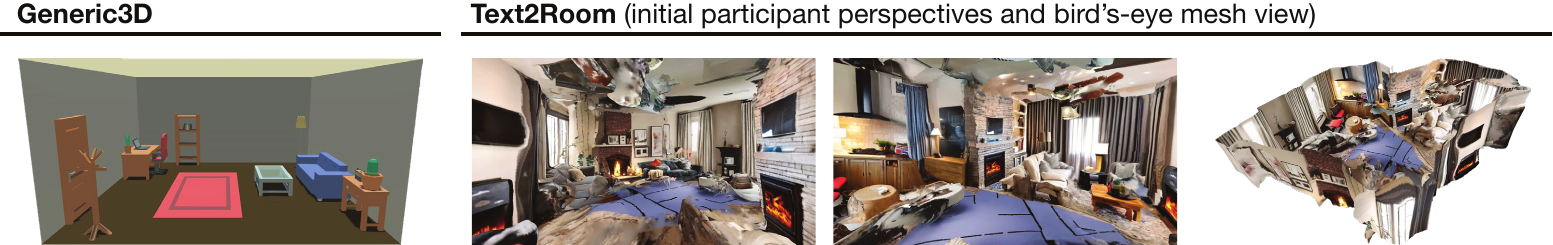}
    \caption{Overview of the environments used for the \GenericCondition and \TextToRoomCondition conditions.}
    \label{fig:condition-comparison}
\end{figure*}

\section{Preliminary User Study}
Our preliminary user study explored the effects of \textit{space blending} in the context of a collaborative affinity diagramming task within a VR telepresence environment.  The study used a within-subjects design with three conditions that emphasized different visual and geometric qualities of virtual environments. With the
selection of conditions below, we sought to explore variations across the dimensions of environmental visual and geometric complexity, fidelity, and familiarity to study their impacts on user behavior and strategies.

\begin{itemize}
\item \textbf{\GenericCondition} served as a baseline representing low-poly environments commonly used in current social VR platforms such as Meta Horizon and Recroom. This space was designed with 3D models from the public domain\footnote{\url{https://kenney.nl/assets/furniture-kit}}.
\item \textbf{\TextToRoomCondition} served as a baseline representing an environment generated with a state-of-the-art 3D scene generation method, containing salient landmarks that contrast with the simple landmarks of the \GenericCondition condition. This environment was produced by the \TextToRoomCondition framework \cite{holleinText2RoomExtractingTextured2023}, which we extended to develop \SystemName.
\item \textbf{\SpaceBlenderCondition} represents our pipeline with participant-provided input images of spaces familiar to them.
\end{itemize}

\begin{figure*}
    \centering
    \includegraphics[width=0.91\textwidth]{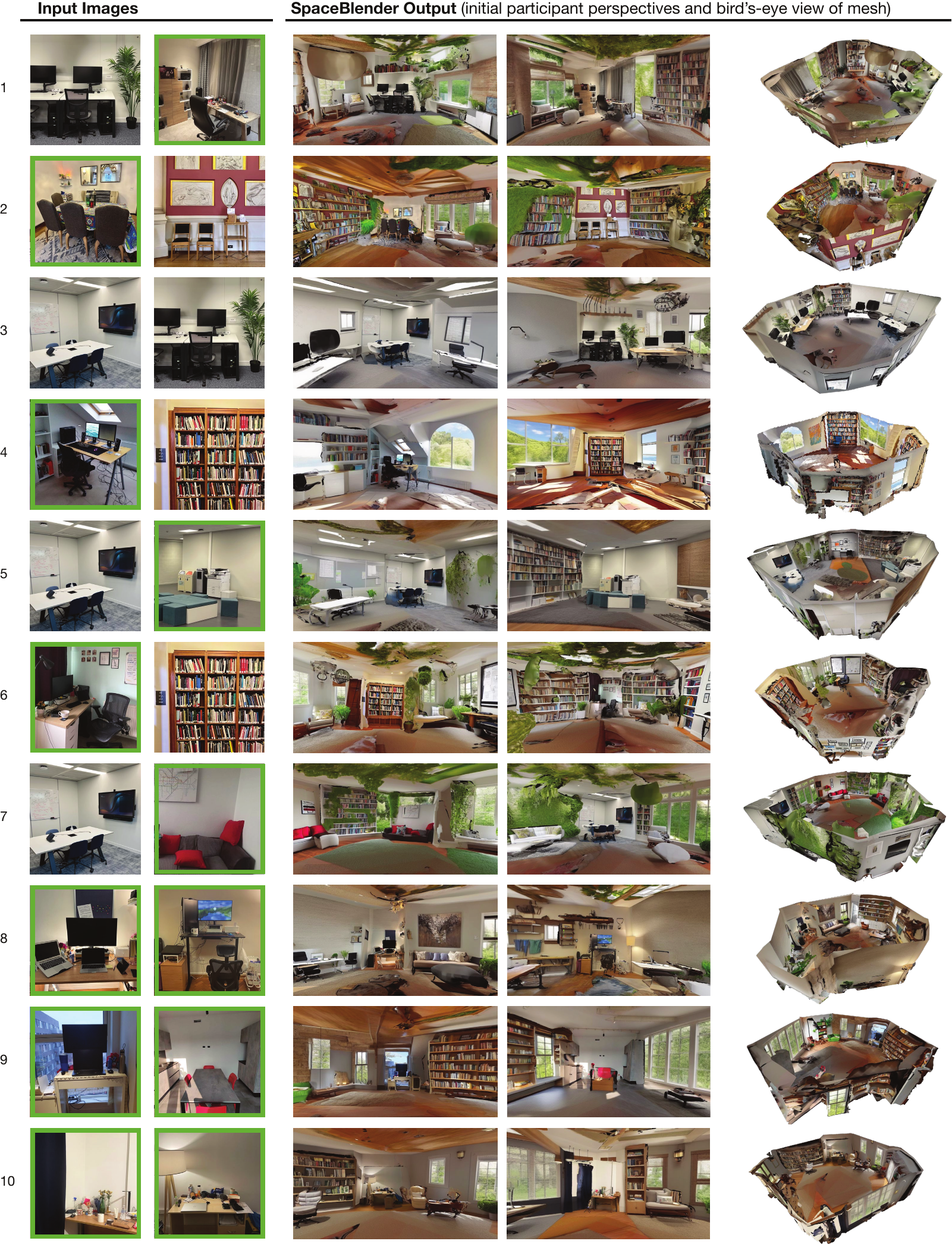}
    \caption{Overview of \SystemName meshes generated based on input images uploaded (with \textcolor[HTML]{70AD47}{green} outline) or selected by participants (no outline).}
    \label{fig:study_mesh_overview}
\end{figure*}

The order of conditions was counterbalanced. The \GenericCondition and \TextToRoomCondition environments were consistent for all pairs and are shown in Fig. \ref{fig:condition-comparison}. For the \TextToRoomCondition condition, we used the pipeline’s public source code and trajectory files to generate twelve environments and then selected the best one based on subjective visual analysis of their geometric and visual quality. For the \SpaceBlenderCondition condition, a new environment was generated for each pair to embed a familiar physical context for each participant, which are shown in Fig.~\ref{fig:study_mesh_overview}. This involved collecting images via an online form sent to participants before the study, where they could upload a photo of a familiar space (e.g., a library, cafe, living room, or desk). Participants who did not upload a photo could choose from seven photographs of various spaces within our institution. Those who found none of the spaces familiar were excluded from the study. Uploaded images were cropped to a 1:1 aspect ratio, excluding any personally identifiable content (e.g., portraits). The study was approved by the University College London Research Ethics Committee (Study ID UCL/CSREC/R/16).

\subsection{Task}
The task involved clustering virtual sticky notes with predefined text. Initially, each participant independently clustered twelve sticky notes from one of two datasets (\textit{fruits} or \textit{vegetables}) by color (e.g., placing “banana” near “lemon”). After two minutes of individual work elapsed, participants were given three more minutes to collaboratively reorganize their clusters into new groups.

\subsection{Recruitment}
We recruited 20 participants (10 pairs) through internal mailing lists. Participants (7 Female, 13 Male) had an average age of 26 years (SD = 6.9) and were mainly students and professionals; seven had backgrounds in computer graphics, and four in UX/HCI. All participants had used a VR headset at least once, with eight using them monthly, and three using VR telepresence platforms monthly. Five pairs knew each other before the study. Participants received £15 gift cards as compensation.

\subsection{Implementation and Setup}
The VR telepresence system for our study was built with Unity3D and the Ubiq framework~\cite{fristonUbiqSystemBuild2021}, providing voice chat, networked objects, and low-poly floating-body avatars. Each participant used a Meta Quest 3 VR headset tethered to a desktop computer with an NVIDIA 4090 RTX GPU in separate physical spaces. Navigation was achieved using the controller’s joystick.

Two stacks of sticky notes were placed in the virtual environments. Participants could grab the top note by pressing and holding the grip button when their virtual hand was near the stack. Given the high complexity of the meshes generated by generative models, sticky notes could be placed anywhere without physics constraints. The stacks were manually placed (e.g., on a table) within arm’s reach of the participant spawn points before the study.

\begin{figure*}
  \includegraphics[width=\textwidth]{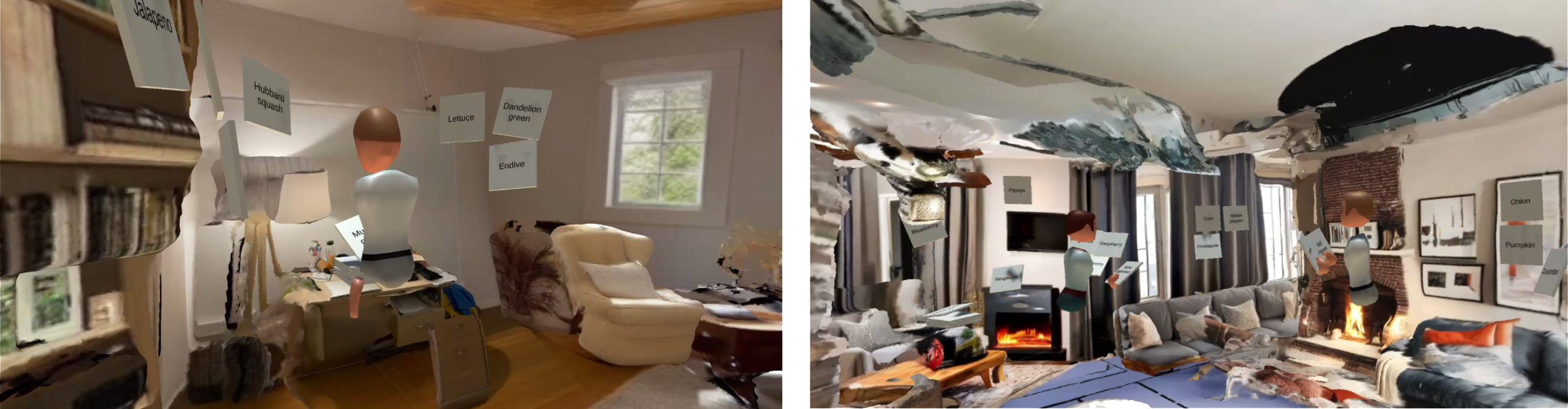}
\caption{Captures of participants manipulating sticky notes while represented by Ubiq avatars in the individual (left) and collaborative (right) task phases.}
\label{fig:participant-screenshots}
\end{figure*}

\subsection{Procedure}

Upon arrival, the experimenter introduced the study’s structure and goals to the participants, including an explanation of the usage of the VR equipment. Participants then read and signed an informed consent form and completed a pre-study questionnaire covering demographics and VR experience. Participants were guided to randomly assigned spaces and equipped with VR headsets. The experimenter then joined the environment from a desktop computer in a separate space, explained the task, and guided participants through a test environment to acclimatize them to the navigation controls and manipulation techniques of the virtual sticky notes.

Participants were then teleported to opposite sides of an environment matching the current condition. In the \SpaceBlenderCondition condition, teleportation points were defined to be at the center of the submesh generated based on the image provided by the respective participant. Participants started by individually clustering the sticky notes. After two minutes, the experimenter instructed them to continue the task collaboratively, combining their individual work for three more minutes (Fig. \ref{fig:participant-screenshots}). After each condition, participants completed a post-task questionnaire.

After the final condition, participants completed a post-study questionnaire. The study concluded with the experimenter guiding participants to a common physical space for a semi-structured interview, including a brief scenario walkthrough. This walkthrough featured two scenarios: a collaborative study session and a cooking class with friends, as shown in Fig. \ref{fig:walkthrough-overview}. The interview questions are shown in ~\ref{appendix:interview-questions}.

\subsection{Data Collection \& Measures}

The post-task questionnaire included questions to measure participants’ perceived spatial presence from an existing questionnaire~\cite{hartmannSpatialPresenceExperience2016}, specifically focusing on two dimensions: Self-Location, which assesses the sensation of being physically present within the virtual environment, and Possible Actions, which measures the perceived capability for interaction within that space. We also incorporated questions to measure perceived Copresence to evaluate participants' perceptions of sharing the virtual environment with others \cite{harmsInternalConsistencyReliability2004}. For each question, "Do not agree at all" was treated as 1, and "Fully Agree" as 5.

Additionally, we included four custom questions to assess how various factors influenced task execution: (1) \textit{layout}, (2) \textit{visual quality}, (3) \textit{level of familiarity}, and (4) \textit{navigation controls}. Each question asked, “To what extent did [$X$] help or hinder you in conducting the task?” with $X$ representing one of the factors. Participants responded using a 5-point Likert scale: \textit{significantly hindered} (1), \textit{slightly hindered}, \textit{neither helped nor hindered}, \textit{slightly helped}, and \textit{significantly helped} (5). We averaged the scores of each measure to arrive at a single score for each.

\section{Results}
In this section, we present \textit{(1)} our quantitative analysis of the participants' self-reported measures; \textit{(2)} qualitative themes around the benefits and limitations of the three environments for the clustering task; and \textit{(3)} participants' suggestions for future requirements and potential use cases of \SystemName's environments.

\begin{figure*}[ht]
    \centering
    \includegraphics[width=\textwidth]{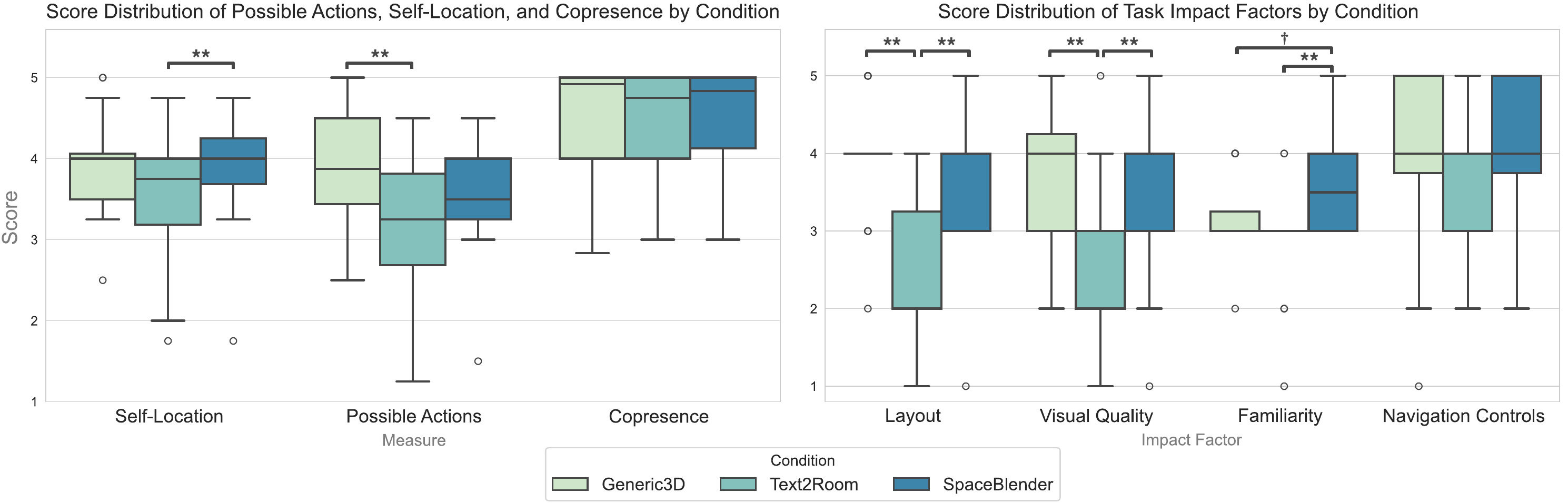}
    \caption{Plot of score distributions of Possible Actions, Self-Location, Copresence, and task impact factors. Levels of statistical significance: \(^\dagger\) for \(p < 0.05\) (before Bonferroni correction), \(^*\) for \(p < 0.05\), \(^{**}\) for \(p < 0.01\), and \(^{***}\) for \(p < 0.001\) (after Bonferroni correction).}
    \label{fig:questionnaire_plot}
\end{figure*}

\begin{figure}
    \centering
    \includegraphics[width=\columnwidth]{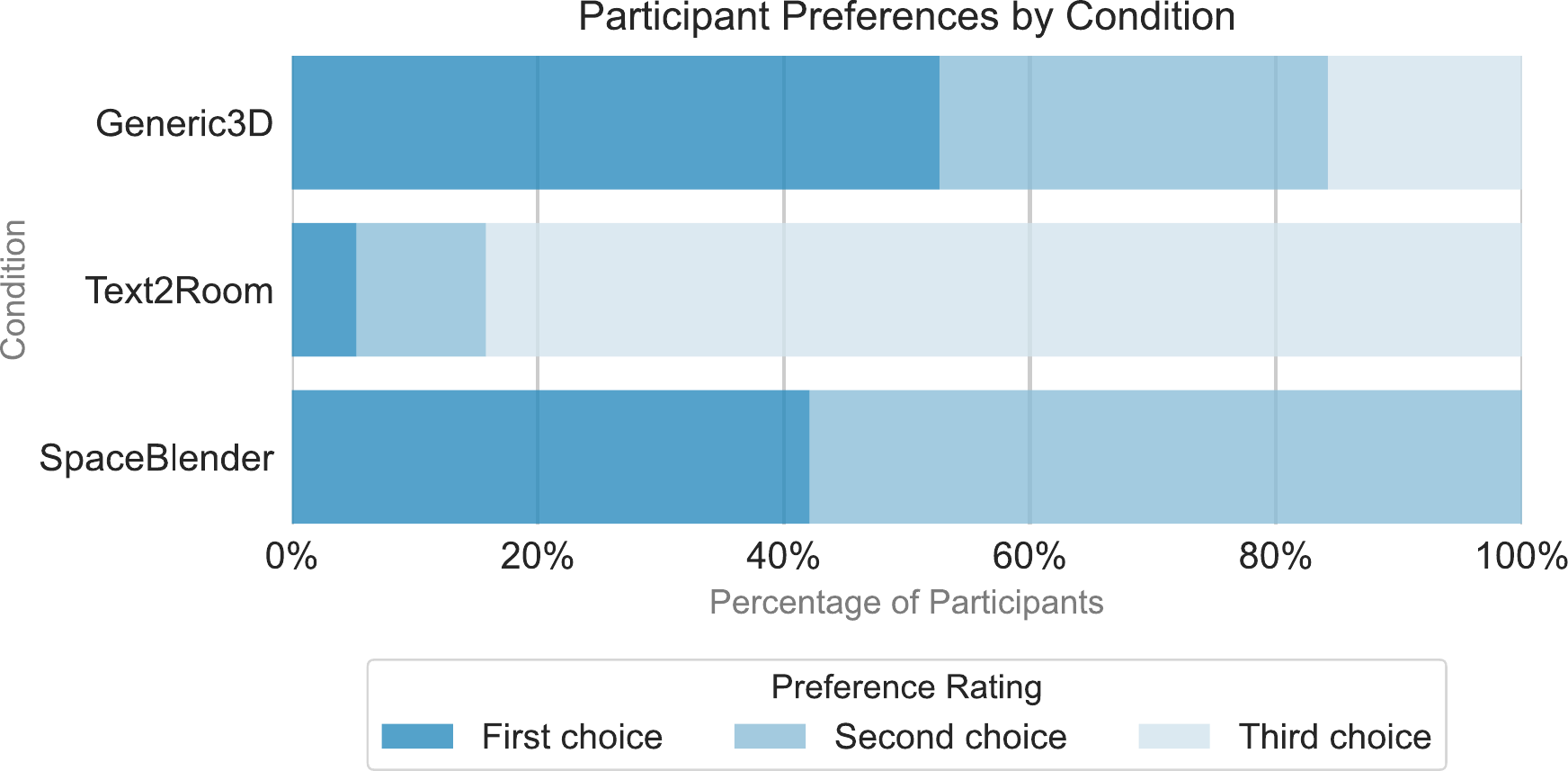}
    \caption{Plot of participant preferences per condition.}
    \label{fig:preference_plot}
\end{figure}

\subsection{Self-Reported Questionnaire Results}

To analyze the self-reported questionnaire data, we first applied the Friedman test to identify overall differences across conditions. Following this, we conducted Wilcoxon signed-rank tests for post-hoc comparisons. Key findings are presented below, with additional results provided in Sec. \ref{appendix:full-results}. Fig.~\ref{fig:questionnaire_plot} shows distributions of participants' scores for Possible Actions, Self-Location, Copresence, and task impact factors. Furthermore, Fig.~\ref{fig:preference_plot} shows participants' ranked preferences for using the \GenericCondition, \TextToRoomCondition, and \SpaceBlenderCondition environments. Most participants ranked the \GenericCondition environment as their first choice for completing the clustering task, followed closely by \SpaceBlenderCondition, which received only first or second choice ratings.

\paragraph{Possible Actions} A statistically significant difference in Possible Actions scores was found between \GenericCondition (M = 3.89, SD = 0.71) and \TextToRoomCondition (M = 3.13, SD = 0.95) ($W = 5.0$, $p = 0.0021$), indicating a diminished perception of possible actions within the environment under the \TextToRoomCondition condition.

\paragraph{Self-Location} A statistically significant difference in Self-Location scores was found between \TextToRoomCondition (M = 3.46, SD = 0.78) and \SpaceBlenderCondition (M = 3.91, SD = 0.68) ($W = 0.0$, $p = 0.0039$), indicating a decreased sense of being physically present within the environment under the \TextToRoomCondition condition.

\paragraph{Impact of Environmental Factors} Among the impact of environmental factors, we found statistically significant differences for the impact of \textit{Layout}, \textit{Visual Quality}, and \textit{Familiarity}. For \textit{Layout} we found statistically significant differences between \GenericCondition (M = 3.92, SD = 0.68) and \TextToRoomCondition (M = 3.20, SD = 0.75) ($W = 10.0$, $p = 0.0043$); and between \TextToRoomCondition and \SpaceBlenderCondition (M = 3.85, SD = 0.70) ($W = 4.0$, $p = 0.0036$). We also found statistically significant differences for \textit{Visual Quality} between conditions \GenericCondition (M = 4.00, SD = 0.00) and \TextToRoomCondition (M = 3.17, SD = 0.75) ($W = 0.0$, $p = 0.0010$); and between \TextToRoomCondition and \SpaceBlenderCondition (M = 3.90, SD = 0.30) ($W = 3.5$, $p = 0.0049$). Lastly, we found statistically significant differences for \textit{Familiarity} between \TextToRoomCondition (M = 3.10, SD = 0.83) and \SpaceBlenderCondition (M = 3.95, SD = 0.22) ($W = 0.0$, $p = 0.0039$).

\subsection{Benefits and Limitations of Environments for the Clustering Task}
Next, we discuss four themes from participants' post-task reflections on the clustering task and to what extent the three styles of virtual environments supported their work.
We refer to paired participants as P\#A and P\#B, where \# represents their pair ID.

\subsubsection{\textbf{Environments typically played a passive role in supporting spatial organization, but some participants adapted their clustering strategies to \SystemName's environments' distinct or familiar features.}}
\label{sec:results:passive-role}
When asked what strategies they adopted to complete the clustering task across all three conditions, a majority of participants described using the center of the environment as a staging area for finalized clusters.
Preferences varied for storing and comparing ungrouped notes either in the middle or in individually assigned regions.

Although the task did not require explicit use of the environment, some participants in \SpaceBlenderCondition utilized its unique features.
First, P5A and P6A expressed that \SpaceBlenderCondition's more detailed environments with distinct segments helped to establish mental models of where to organize notes:
``I just put the green objects in the green area of the environment'' (P6A).
This strategy contrasts with P6A's experience in \GenericCondition, where a lack of environmental cues led to a more deliberate strategy of labeling areas of the environment to place specific clusters: ``We had to actually allocate areas because the wall was just gray.''
Additionally, several participants found value in the familiar details of \SpaceBlenderCondition's environments (i.e., spatial landmarks preserved from their input images) to inform their clustering strategies (P4A, P7A, P10A).
P7A noted: ``Familiarity helped because this is pretty much like where I work a lot''; ``it just feels a bit more like comfortable, like, thinking in that area.'' 
This sentiment was echoed by P10A, who envisioned aligning sticky notes to their own table in the virtual workspace.

\subsubsection{\textbf{Participants had mixed preferences for minimalistic and realistic environments for supporting their focus on the clustering task.}}
\label{sec:results:fidelity}
A majority of participants favored the simplistic design of \GenericCondition as they believed it enabled them to be more engaged with the task. 
P8A felt that as the ``the cleanest environment,'' the \GenericCondition environment supported task efficiency.
Although P3A perceived the \GenericCondition environment to be ``cartoonish'' and ``not real,'' they considered it the ``highest quality'' environment. 
P9A identified benefits to having both low-fidelity environments and virtual avatars:``I know my figure in the space was artificial, so I'm more aware I'm in the game.''

However, not all participants preferred the minimalistic environments: ``[\textsc{Generic3D}] is my least favorite, because it didn't feel realistic'' (P4B).
Some participants found value in working in the more detailed \TextToRoomCondition and \SpaceBlenderCondition environments: ``because it feels more realistic, it feels more immersive,''  despite the potential for these details to detract from their task engagement.

\subsubsection{\textbf{Participants perceived increased physical comfort and navigability in the \SpaceBlenderCondition and \GenericCondition conditions, as opposed to the \TextToRoomCondition condition.}}
All participants noted \textsc{Text2Room}’s inconsistent floor geometry as a major source of navigational difficulty and discomfort, with P8A noting that ``stuff sticking out of the floor'' significantly hindered their ability to navigate in the space. 
Similarly, P3B pointed to ``noise and also the distortion in the floor'' in \TextToRoomCondition, but noted that this issue was not present in \SpaceBlenderCondition. 

However, a majority of participants highlighted that the both generative environments exhibited low texture resolution, noisy artifacts, and incoherent geometry, which could detract from the realism and usability of the spaces.
Both the \TextToRoomCondition and \SpaceBlenderCondition environments were criticized for their geometric inaccuracies that led to physical objects like ``the table and a couple of the chair legs'' being mapped onto the floor (P5B).

At times, the physical discomfort reported by participants extended beyond visual annoyance. 
For instance, P10A stated that they ``did not want to move much in [\textsc{Text2Room}],'' attributing their simulator sickness to the environment's poor construction and excessive clutter. 
The impact of these challenges was so pronounced that some participants stated having to close their eyes while navigating in \TextToRoomCondition spaces (P5B, P10A, P11A, P11B).

In contrast, the \GenericCondition environments' visual and geometric consistency were universally praised for being ``navigable'' and ``clean''. 
Participants found the simplistic and stylized nature of the \GenericCondition environment not only visually appealing but also conducive to task performance. 
P6A appreciated the environment being ``more spacious and a lot more easy to navigate.'' 
Similarly, P8A described their experience in the \GenericCondition environment as ``clean and easy to perform the task in.''

\subsection{Future Requirements and Potential Use Cases for Blended Spaces}
\label{sec:feedback-implementation-applications}
Finally, we discuss participants' comments regarding future scenarios where \SystemName environments could benefit VR telepresence and suggestions for improvement of the blended environments.

\subsubsection{\textbf{Participants envisioned a variety of future scenarios where blending familiar context into virtual environments could explicitly or implicitly provide value.}}
\label{sec:res:future-scenarios}
After walking through additional \SystemName environments, participants brainstormed how \SystemName may be applied within VR telepresence environments in the future.
For professional use cases, participants were interested in meeting spaces incorporating inspiring physical locations (e.g., ``areas where you can have a coffee... and discussions'') to be ``more conducive for ideation'' (P6B) and learning environments that use different parts of the blended space to structure educational activities (P3A, P5A, P6B). 
They also saw value in supporting social interactions (e.g., gaming with friends, family gatherings) and personal well-being (e.g., providing familiar environments for therapy and recording personal memories) (P3A, P3B, P4A P8A).
Some participants expressed concerns regarding the exposure of personal spaces (P3A, P3B, P8A, P7A). For instance, P3A expressed that depending on the scenario, they may feel uncomfortable sharing “a part of [their] life”.

Reflecting on these scenarios, participants distinguished between \SystemName environments providing \textit{explicit} and \textit{implicit} value for collaboration in different contexts. 
For example, virtual training applications may need realistic depictions of users' surroundings, while social use cases could benefit indirectly from familiar details that ``invoke a sense of being in a cozier space'' (P8A).
This distinction aligns with participants' feedback on the clustering task: some felt the \SpaceBlenderCondition environment did not explicitly support their clustering strategies, while others described implicitly using environmental features (e.g., colors, familiar furniture) to establish organization patterns.

\subsubsection{\textbf{Enabling future use cases for blended environments in VR telepresence requires quality improvements and real-world alignment.}}
\label{sec:res:future-usecases}
While participants could see the benefits of using \SystemName environments in VR telepresence systems, participants universally emphasized the need to improve visual quality and realism to fully realize these benefits.
First, we observed instances where environments deviating from participants' memories of familiar locations caused confusion or disorientation. 
P9A expressed concerns about the accuracy of input images reconstructed in the blended space: ``it made me feel a bit weird that I could not recognize some things on my desk.''
P1A mentioned needing to adjust expectations due to partial reconstructions of familiar environments: ``it takes a lot of getting used to if there’s... somewhere you're familiar with, but then you turn around and it's not what you're familiar with.''
Some participants also noted that the shape of the blended spaces did not match their expectations (P1B, P1A): "[…] closed spaces usually don't look like that" (P1B), while P4A suggested a specific use case for the circular shape of the blended space, imagining "[...] students in a circle, and the teacher standing in the middle."

To improve realism, some participants wanted the blended virtual environment to ``connect with [their] real environment'' by matching the geometry of their physical surroundings (P9B).
With this approach, participants envisioned enabling mixed reality collaboration for physical tasks (P1B, P3A, P10A, P10B, P11B), such as in our cooking class scenario (Fig. \ref{fig:walkthrough-overview}).

\section{Discussion and Future Work}
We structure our discussion around \textit{(1)} avenues for further studies to explore the impact of blended virtual environments on VR collaborative tasks; 
\textit{(2)} themes for potential improvement and extension of \SystemName;
\textit{(3)} the limitations of our work.

\subsection{Opportunities for Supporting VR Telepresence with Blended Environments}

\subsubsection{\textbf{Leveraging familiar context in VR telepresence scenarios}}

Several participants noted that familiar elements in the \SpaceBlenderCondition environment supported their clustering strategies by providing contextual cues similar to their real-world experiences. Similarly, questionnaire responses indicated that familiarity had a greater impact on the clustering task in the \SpaceBlenderCondition condition compared to the other two conditions. Although our study was preliminary and requires further empirical research, these early insights suggest that familiarity is a promising direction for future research on collaborative spaces.

Participant feedback indicated that the use and desired functionality of blended spaces may vary between professional and social contexts. In professional settings, blended spaces might be used to establish workspace context and create environments that stimulate creativity. With continued usage, we envision that familiar spaces could be tools for preparation and augmenting memory for collaborative tasks in blended spaces. Furthermore, after collaboration, the uniquely blended regions of the \SystemName spaces could provide a lasting artifact of the collaborative work~\cite{rajaramBlendScapeEnablingEndUser2024}. In social contexts, familiarity may play a more implicit role, recreating comfortable and meaningful environments. Future research could explore how blending familiar personal spaces affects social presence and personal identity.

Despite the potential of blending, some participants expressed discomfort with merging familiar and novel spaces, highlighting the need for careful consideration of privacy and personal boundaries. Future research could investigate mechanisms that allow users to control the degree of blending and ensure only intended elements of their familiar spaces are shared.

\begin{figure}
    \centering
    \includegraphics[width=\columnwidth]{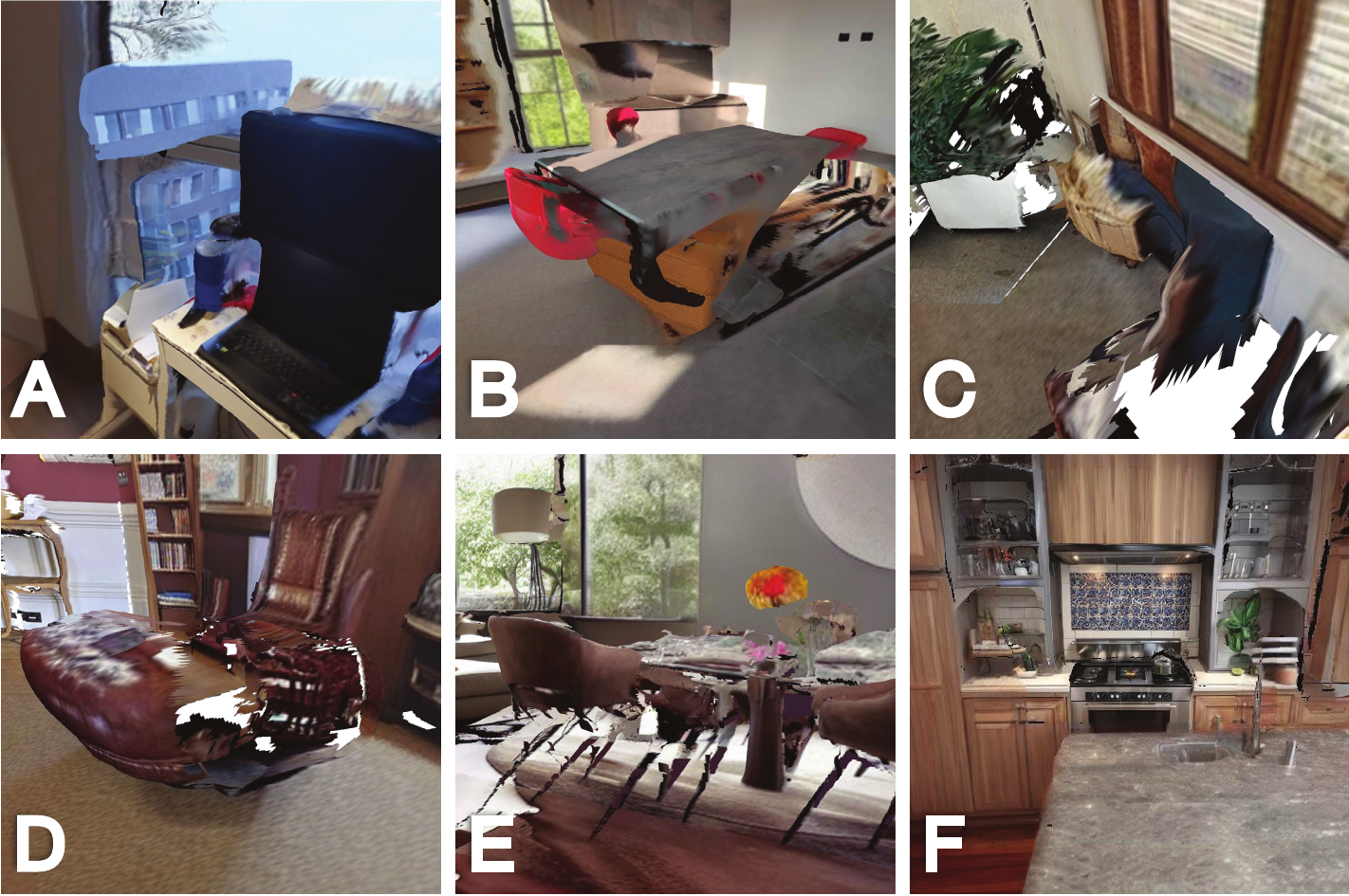}
    \caption{Examples of \SystemName pipeline components errors: (A) protrusion of outdoor structure, (B) distorted table, (C) depth estimation error, (D) depth alignment error, (C) depth estimation error, (F) floor expansion error.}
    \label{fig:failure_overview}
\end{figure}

\subsubsection{\textbf{Extending \SystemName environments to support explicit interactions for collaborative scenarios.}}
The \SystemName environments primarily implicitly supported the clustering task in our user study by grounding users in familiar spaces. We envision several ways to extend \SystemName to explicitly support collaboration. One such extension could enable users to manipulate virtual objects within the scene by semantically segmenting scene components and applying matching functionalities, such as \textit{drawing} on a \textit{whiteboard}~\cite{delatorreLLMRRealtimePrompting2024,giunchiDreamCodeVRDemocratizingBehavior}. In line with this, while generating 3D scenes is currently time-consuming, we foresee future models allowing for real-time user-driven changes to customize spaces, similar to those available in 2D generative systems, with future studies possibly exploring customization preferences~\cite{rajaramBlendScapeEnablingEndUser2024}.

Future work may also explore extending the layout, geometric prior, and trajectory definition techniques to support more ecologically valid submesh arrangements, including multi-room or multi-story structures, enabling features like breakout rooms and meeting context transitions~\cite{gonzalezdiazMakingSpaceSocial2022}. To further incorporate spatial familiarity, these layouts could be modeled after existing building floor plans, with submeshes aligned to these layouts instead of the current parameter-based layout approach (Sec.~\ref{sec:system:layout}). Future work may study how these spatially familiar layouts impact user navigation, collaboration efficiency, and spatial awareness in virtual environments.

\subsection{Opportunities to Improve and Extend \SystemName}
In this section, we discuss potential ways to improve and extend \SystemName based on feedback shared by our study participants.

\subsubsection{\textbf{Improving the quality and physical comfort of \SystemName environments.}}

Participants highlighted the need for enhanced visuals and geometry to support their envisioned future uses of \SystemName. A prevalent issue associated with this feedback was inaccurate depth estimation and alignment, leading to objects merging with walls, floors, or ceilings, or displaying implausible depth values, diminishing the realism and coherence of the scene (Fig.~\ref{fig:failure_overview}A–E). Due to the iterative nature of the pipeline and the dual use of the mesh representation for rendering and completion, these inaccuracies tended to amplify during generation. One way to address this may be to replace IronDepth with a more recent monocular depth estimation model~\cite{bhatZoeDepthZeroshotTransfer2023,yangDepthAnythingUnleashing2024}. However, the replacement model should also support depth image \textit{inpainting}, a task that is less commonly supported. Additionally, semantic priors could be extended to objects to improve geometric consistency, as used by the recent framework ControlRoom3D~\cite{schultControlRoom3DRoomGeneration2024}. While this framework requires the manual definition of these priors, it could be combined with our geometric prior definition technique and existing furniture layout synthesis methods for an automated approach~\cite{paschalidouATISSAutoregressiveTransformers2021,fengLayoutGPTCompositionalVisual2023}. Alternatively, SDF-based methods may be used to define the mesh directly, which could be extended to make the generation conditional on the geometry of input submeshes to support blending.

Furthermore, several participants noted the low resolution of \SystemName environments. While the current MultiDiffusion-based inpainting process would support higher resolution images if paired with depth estimation and inpainting model capable of processing these, adopting more recent image generation models that produce higher-resolution images could further improve visual quality~\cite{sauerFastHighResolutionImage2024}. Some recent models~\cite{liGligenOpensetGrounded2023,phungGroundedTexttoImageSynthesis2024} also offer improved controllability, which could help prevent failure modes such as the one in Fig.~\ref{fig:failure_overview}F, where our floor generation method produced a kitchen island instead of a floor, even though the prompt specified otherwise. Lastly, recent video generation models~\cite{wangMotionCtrlUnifiedFlexible2023,dengStreetscapesLargescaleConsistent2024} might replace iterative image inpainting to enhance visual quality through multi-view consistency without relying on a mesh as an intermediate representation.

\subsubsection{\textbf{Aligning \SystemName environments with the real world.}}
Our participants commonly expressed a desire to align the blended space with their physical environment (Sec. \ref{sec:res:future-usecases}). This may be achieved by extending \SystemName to accept meshes or point clouds as input pre-captured by users or, alternatively, captured by RGB-D cameras mounted in the user’s local space in real-time~\cite{lindlbauerRemixedRealityManipulating2018}. These representations could be registered to the user's local physical space, who could then use a mixed-reality headset to observe the blended space.

\subsection{Limitations}

Our user study included both VR novices and periodic VR users, whose perceptions may not generalize to experienced VR users. However, we note that our novice VR users helped surface the most critical challenges with \SystemName's environments, particularly those involving core VR interaction requirements (e.g., physical comfort and navigability) that pertain to experienced users as well.

Furthermore, some elements of our study design limited our ability to isolate the impacts of familiar context and fidelity on participants' task performance and collaboration patterns.
First, we did not control the familiarity of participants’ input images to generate \SystemName environments, as some users used pre-selected images instead of uploading their own.
Second, our choice of baseline conditions may limit the generalizability of some of our findings. A number of participants preferred the higher realism of the generative environments over the \GenericCondition condition. However, since the baseline and generative conditions had vastly different types of texture quality, this participant preference for the generative conditions might be different if compared to a higher-fidelity version of the \GenericCondition baseline.
Future work could compare \SystemName with other environment types, including such higher-fidelity virtual environments\footnote{E.g., \textit{Spatial}: \url{https://spatial.io}, \textit{Microsoft Mesh}: \url{https://microsoft.com/mesh}}, 3D scans, or manually designed environments.
Third, the clustering task did not require participants to interact with the virtual environment explicitly, allowing them to choose whether to use environmental features for the spatial organization of sticky notes. We chose affinity diagramming as a fair task across all three conditions because the \GenericCondition and \TextToRoomCondition environments would be insufficient for tasks requiring access to users’ physical surroundings. Future work may explore studying blended spaces combined with VR tasks that require explicit environmental interaction. Lastly, considering the novelty of generative 3D environments incorporating familiar environments, participant responses may have been subject to response bias~\cite{dellYoursBetterParticipant2012}.

Additionally, we acknowledge that our sample size was insufficient to reliably calculate order effects. Participants who began with the \GenericCondition might have been primed to organize notes in mid-air rather than aligning them with the environment in subsequent conditions due to the environment’s lack of distinctive visual features.

\section{Conclusion}
To enable the creation of context-rich virtual spaces for VR telepresence, our work contributes \SystemName, a pipeline that leverages generative AI to incorporate and extend users' physical surroundings into blended virtual environments.
\SystemName makes key improvements to current state-of-the-art generative models by projecting multiple user-provided images into 3D segments, aligning mesh segments to a uniform floor level, and blending those segments via diffusion-based space completion methods guided by geometric priors and dynamic text prompts.
Through a preliminary within-subjects study with 20 participants, we explored how varying the virtual environment (using \GenericCondition, \TextToRoomCondition, and \SystemName environments) affects their behavior and strategies when completing a collaborative clustering task.
Overall, participants experienced increased physical comfort and navigability in the \GenericCondition and \SpaceBlenderCondition compared to \TextToRoom due to greater consistency in the room geometry. Furthermore, some leveraged recognizable environmental features in the \SpaceBlenderCondition space to complete the task.
Additionally, participants envisioned a rich set of professional, social, and personal use cases where embedding familiar contextual details into virtual environments could provide value for collaboration. 
However, to fully realize the potential benefits, they desired further aligning \SystemName environments to real-world spaces and enhancing their visual and geometric quality. 

Given the current gap in the HCI community's understanding of deploying generative AI environments in interactive systems, our studies around \SystemName lay the groundwork for future generative AI-based systems for VR environment creation. 
We note promising avenues for future work to extend and deploy our pipeline in VR telepresence systems to further study the impact of blended environments on collaborative processes.

\begin{acks}
We thank Anthony Steed for providing valuable feedback and suggestions on our study design. We also extend our gratitude to the reviewers for their insightful comments and our study participants for their time. The preliminary user study of this research was partially supported by the European Union’s Horizon 2020 Research and Innovation program under grant agreement No. 739578.
\end{acks}

\bibliographystyle{ACM-Reference-Format}
\bibliography{ZoteroReferences, References}

\appendix
\onecolumn
\appendix
\section{Appendix}

\renewcommand\thefigure{\thesection.\arabic{figure}}
\setcounter{figure}{0}
\renewcommand{\thetable}{A.\arabic{table}} \setcounter{table}{0}

\label{sec:appendix}

\subsection{Contextually Adaptive Prompt Inference: Full System Prompt}
\label{appendix:system_prompt}
\paragraph{System Prompt} \texttt{You are a helpful assistant that acts like highly creative interior architect and photographer. You are given descriptions of images captured by a camera on a tripod placed exactly in the middle of an open indoor space at a height of 1.5 meters. This camera has a field-of-view of 55 degrees and only takes square images. You will be given a set of image descriptions with Y rotation values of the camera, as well as Y rotation values without a description. From your perspective as a creative interior architect, your task is to describe what you expect to see for the image that will be taken at the Y rotation value with unknown contents. You receive the rotation value and descriptions in JSON format. Do not use words such as 'blend, 'transition', 'fusion', 'mix', 'transformation' (or synonyms), but concretely describe the novel contents and salient objects you expect to see in the area without repeating objects. Use a similar format to the other given descriptions. It is not always obvious what the camera will capture as the contents of the space can be highly diverse in style and content, so please be creative and focus on coming up with new objects and artifacts that fit in. You can assume that the objects appearing in the known image description do not show up in the other image (all objects are fully contained within the image frame). Do not include mentions of the shape of the room (e.g., corner). Use comma-separated descriptions (e.g., instead of 'On the sticky note wall, a whiteboard marker tray holds colorful pens, a spark of color in an otherwise monochrome environment.', write 'Sticky note wall with whiteboard marker holding colorful pens, monochrome environment'. Always start descriptions with '... space with' where '...' is the type of the room (e.g., living room, kitchen, etc.). Only return the descriptions with the \textit{set\_description} function, without any explanation. Keep your descriptions short please, without adding too many different items/objects, with 20 words or less for each description. }

\paragraph{User Prompt} \texttt{The size of the room is space\_size\_str (WxHxL) meters and the camera is positioned in the middle. These are the Y rotation values and descriptions of the images that were already taken: \textit{y\_rotations\_and\_descriptions}. What do you expect for the following Y rotation values: \textit{y\_rotations\_without\_descriptions}? Consider the theme of "" when coming up with the descriptions}

\newpage
\subsection{ControlNet-Layout: Additional Output Samples}
\begin{figure*}[ht]
    \centering
    \includegraphics[width=\textwidth]{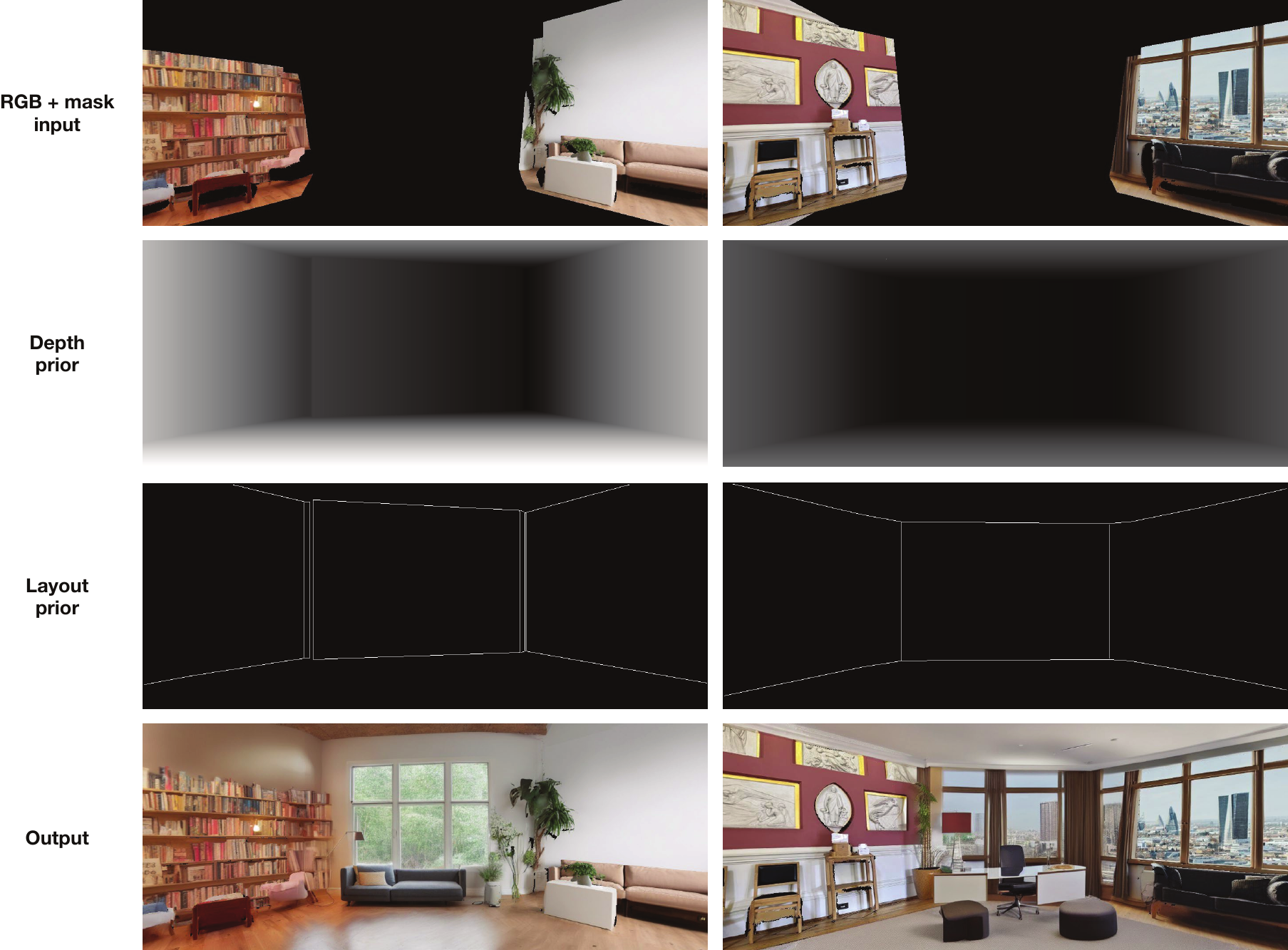}
    \caption{Additional results of the ControlNet-Layout model. Each of these output images was generated by combining Control-Layout and ControlNet-Depth with weights 0.6 and 0.3, respectively.}
    \label{fig:control_layout_extra_samples}
\end{figure*}

\newpage
\begin{figure*}[ht]
    \centering
    \includegraphics[width=\textwidth]{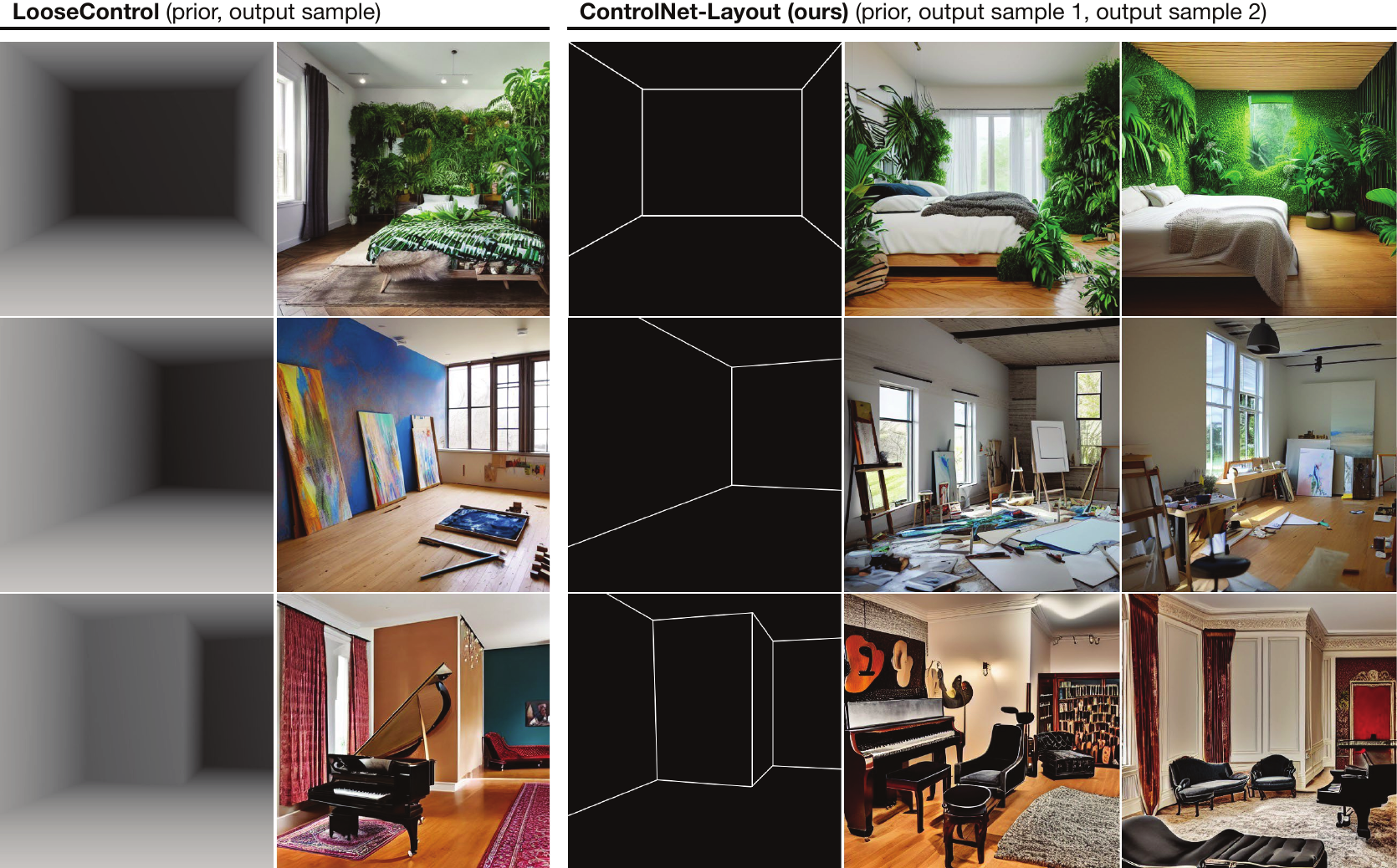}
    \caption{Comparative image generation output of the recent LooseControl model and our ControlNet-Layout model, including prior images and output images each. The LooseControl prior and output images in this figure were reproduced from the paper's web page (\url{https://shariqfarooq123.github.io/loose-control}) with permission from the authors. The ControlNet-Layout prior images were manually created to match the room structure depicted in the LooseControl prior images.}
    \label{fig:controlnet_layout_loosecontrol_comparison}
\end{figure*}

\newpage
\subsection{Influence of Submesh Shapes on Geometric Prior Shape}

\begin{figure*}[ht]
    \centering
    \includegraphics[width=0.9\textwidth]{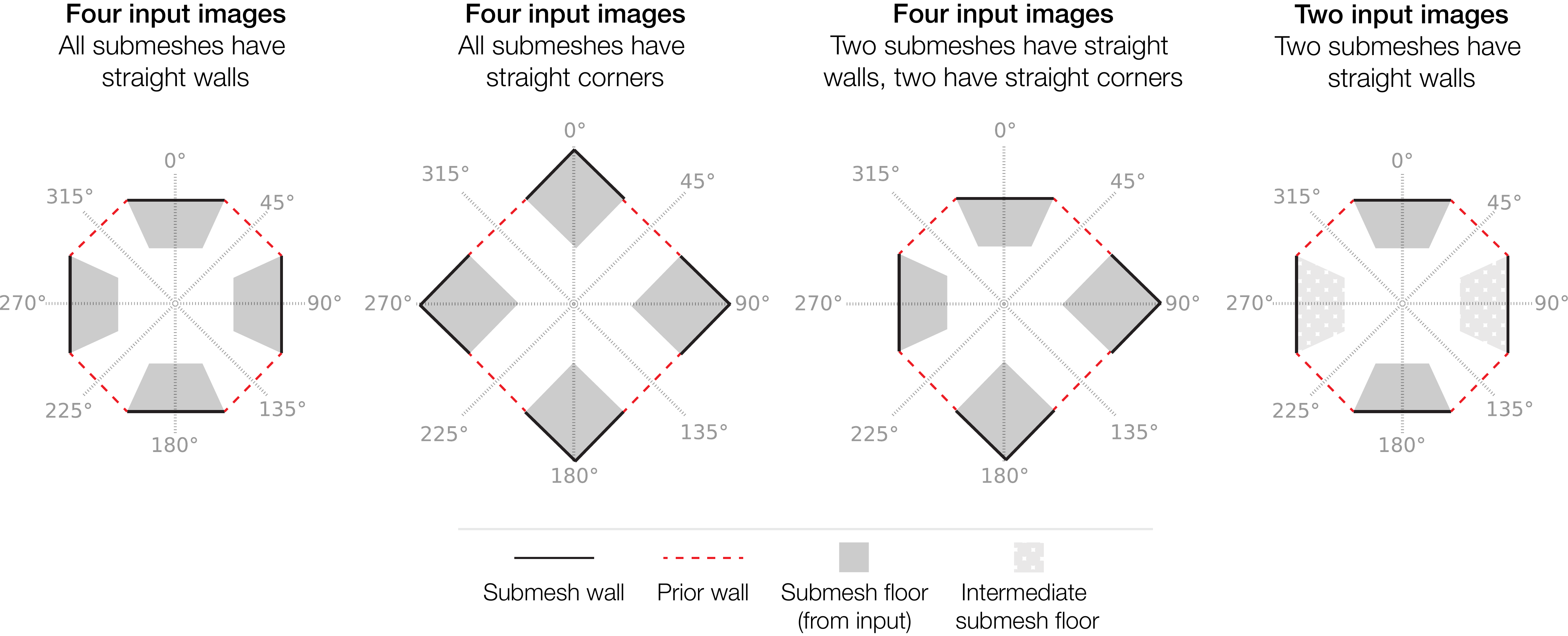}
    \caption{Visual explanation of the impact of submesh count and shape on the shape of the geometric prior mesh and final blended space.}
    \label{fig:prior-influence-explanation}
\end{figure*}

\begin{figure*}[ht]
    \centering
    \includegraphics[width=0.9\textwidth]{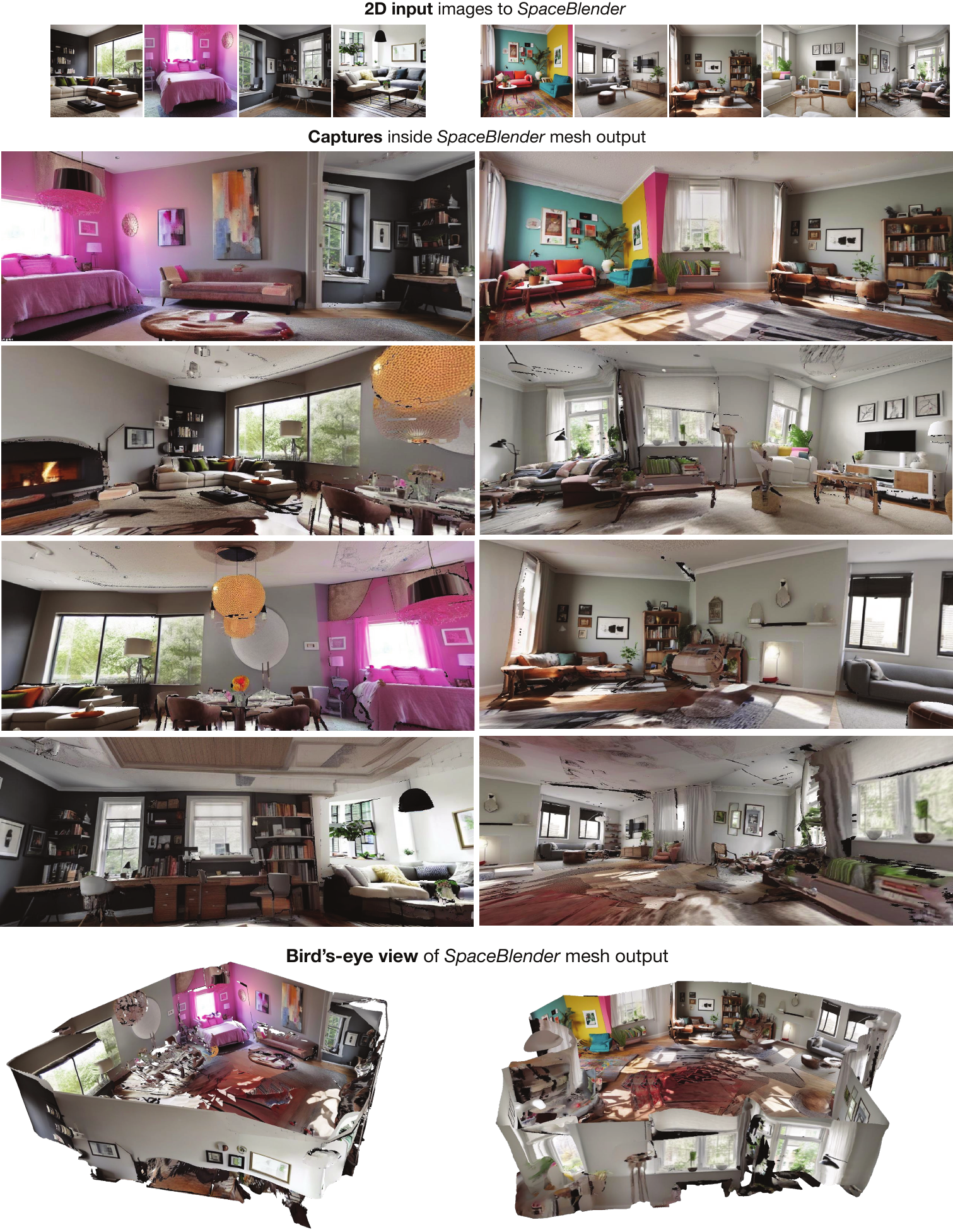}
    \caption{Examples of spaces generated with various numbers of submeshes. Left: a \SystemName mesh based on four input images, all featuring corners. Right: a \SystemName mesh based on five input images featuring a mixture of room shapes.}
    \label{fig:prior-influence-examples}
\end{figure*}

\clearpage
\newpage
\subsection{Preliminary User Study: Questionnaire Items}
\begin{table*}[ht]
\centering
\begin{tabular}{p{0.8\linewidth}l}
\toprule
\textbf{Please indicate your level of agreement with the following statements} & \textbf{Category} \\
Do not agree at all; Disagree; Neutral; Agree; Fully Agree &  \\
\midrule
I felt like I was actually there in the environment of the presentation. & Self-Location \\
It seemed as though I actually took part in the action of the presentation. & Self-Location \\
It was as though my true location had shifted into the environment in the presentation. & Self-Location \\
I felt as though I was physically present in the environment of the presentation. & Self-Location \\
\addlinespace 
The objects in the presentation gave me the feeling that I could do things with them. & Possible Actions \\
I had the impression that I could be active in the environment of the presentation. & Possible Actions \\
I felt like I could move around among the objects in the presentation. & Possible Actions \\
It seemed to me that I could do whatever I wanted in the environment of the presentation. & Possible Actions \\
\addlinespace 
I noticed (my partner). & Co-presence \\
(My partner) noticed me. & Co-presence \\
(My partner's) presence was obvious to me. & Co-presence \\
My presence was obvious to (my partner). & Co-presence \\
(My partner) caught my attention. & Co-presence \\
I caught (my partner's) attention. & Co-presence \\
\bottomrule
\end{tabular}
\caption{Post-task questionnaire for measuring Self-Location and Possible Actions from The Spatial Presence Experience Scale questionnaire by \citet{hartmannSpatialPresenceExperience2016} and Co-presence from the Networked Minds Measure of Social Presence questionnaire by \citet{harmsInternalConsistencyReliability2004}.}
\label{tab:slpa_survey_questions}
\end{table*}

\newpage
\subsection{Preliminary User Study: Walkthrough Scenarios}

\begin{figure*}[ht]
    \centering
    \includegraphics[width=0.85\textwidth]{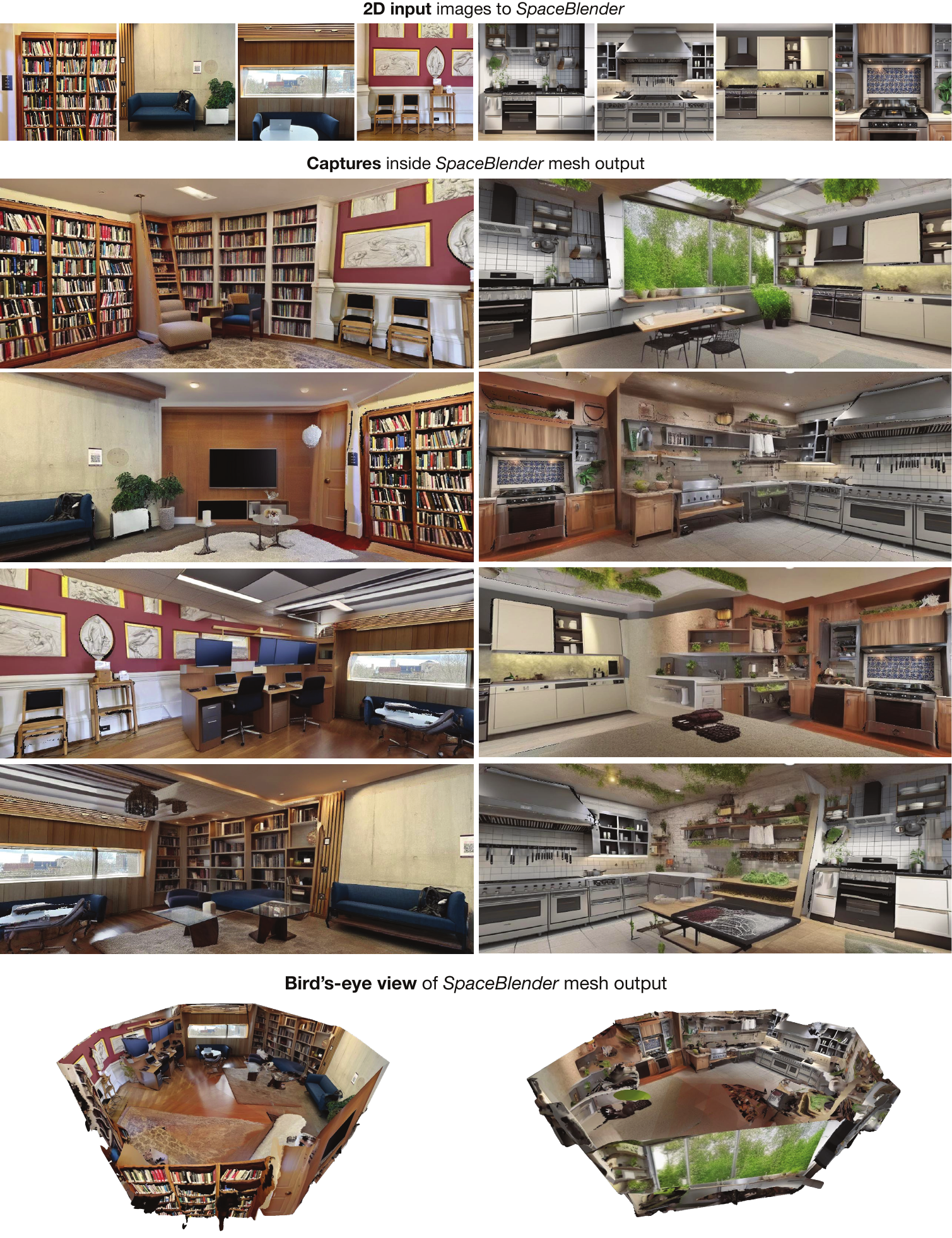}
    \caption{Overview of environments used in the walkthrough segment of the semi-structured interview, including a \textit{collaborative study session} scenario (left) and a  \textit{cooking class with friends} scenario (right).}
    \label{fig:walkthrough-overview}
\end{figure*}

\newpage
\subsection{Preliminary User Study: Semi-Structured Interview Questions}
\label{appendix:interview-questions}
\textbf{Questions involving clustering task:}
\begin{enumerate}
    \item You mentioned you preferred *condition*, over the others. Could you elaborate on your preference?
    \item Describe the strategies you used to perform the affinity diagramming task. Did you adopt a consistent strategy in all three environments, or did you use different strategies?
\end{enumerate}

\textbf{Questions involving walkthrough of additional \SystemName environments:}
\begin{enumerate}
    \item If any, in what scenarios do you think integrating environmental context that you have a personal relation to in a virtual environment could be valuable or interesting?
    \item Is there anything you wished was different about the design, composition, or function of the blended spaces that you’ve seen that could make it more engaging, useful, or comfortable?
\end{enumerate}

\subsection{Preliminary User Study: Full Self-Reported Questionnaire Response Analysis}
\label{appendix:full-results}
We used the Friedman test for overall comparisons for our analysis, with subsequent post hoc analyses conducted via Wilcoxon signed-rank tests. All reported p-values for post hoc analyses have been adjusted using the Bonferroni correction.

Given that the assumptions requisite for parametric tests were not met for a majority of the results, and given the general recommendation to use non-parametric tests in studies with limited sample sizes, our analysis utilized the Friedman test for overall comparisons, with subsequent post hoc analyses conducted via Wilcoxon signed-rank tests. All reported p-values for post hoc analyses have been adjusted using the Bonferroni correction.

\subsubsection{Possible Actions}
The Friedman test revealed a statistically significant difference in the scores of Possible Actions across the three conditions ($\chi^2(2) = 14.81, p < 0.0001$), suggesting that participants' perceptions of possible actions varied significantly depending on the condition. Post hoc analyses were conducted to explore these differences further. Results showed a significant decrease (W = 5.0, $p = 0.0021$) in Possible Actions scores from \GenericCondition (M = 3.89, SD = 0.71) to \TextToRoomCondition (M = 3.13, SD = 0.95), indicating a diminished perception of possible actions within the environment under \TextToRoomCondition. However, the difference between \GenericCondition and \SystemName (W = 23.0, $p = 0.104$) and between \TextToRoomCondition and \SystemName (W = 27.0, $p = 0.055$) did not reach statistical significance.

\subsubsection{Self-Location}
A similar analysis was conducted for Self-Location scores, with the Friedman test indicating a significant difference across conditions ($\chi^2(2) = 10.53, p = 0.0052$). This finding highlights that the different environments significantly affected the sense of being situated within the environment. After post hoc analysis, the comparison between \GenericCondition and \TextToRoomCondition and between \GenericCondition and \SpaceBlenderCondition did not show significant differences (W = 49.0, $p > 0.999$). However, a significant difference (W = 0.0, $p = 0.0039$) was found between \TextToRoomCondition (M = 3.46, SD = 0.78) and \SpaceBlenderCondition (M = 3.91, SD = 0.68), indicating a change in the sense of self-location between these two conditions.

\subsubsection{Co-Presence}
The analysis of co-presence scores using the Friedman test did not reveal a statistically significant difference across conditions ($\chi^2(2) = 1.59, p = 0.452$).

\subsubsection{Impact of Environmental Factors}
The assessment of environmental factors on task performance revealed several statistically significant differences across conditions.

The analysis for \textit{Layout} showed a $\chi^2(2) = 20.22, p < 0.0001$, indicating significant variations in task conductance related to environmental layout. Post-hoc analysis indicated significant differences between \GenericCondition (M = 3.92, SD = 0.68) and \TextToRoomCondition (M = 3.20, SD = 0.75) (W = 10.0, $p = 0.0043$). A significant difference was also observed between \TextToRoomCondition and \SpaceBlenderCondition (M = 3.85, SD = 0.70) (W = 4.0 and $p = 0.0036$). However, the difference between \GenericCondition and \SpaceBlenderCondition did not reach statistical significance (W = 22.0, $p = 0.4842$).

For \textit{Visual Quality}, a significant difference was found ($\chi^2(2) = 21.73, p < 0.0001$). There was a difference in perceived support for task execution between \GenericCondition (M = 4.00, SD = 0.00) and \TextToRoomCondition (M = 3.17, SD = 0.75) (W = 0.0 and $p = 0.0010$), and between \TextToRoomCondition and \SpaceBlenderCondition (M = 3.90, SD = 0.30) (W = 3.5 and $p = 0.0049$). The comparison between \GenericCondition and \SpaceBlenderCondition did not show a significant difference (W = 18.0, $p = 0.2121$).

\textit{Familiarity} also demonstrated significant differences among conditions ($\chi^2(2) = 14.56, p = 0.00069$). The difference between \TextToRoomCondition (M = 3.10, SD = 0.83) and \SpaceBlenderCondition (M = 3.95, SD = 0.22) was significant (W = 0.0 and $p = 0.0039$). Comparisons between \GenericCondition and \TextToRoomCondition, and between \GenericCondition and \SpaceBlenderCondition did not achieve significance (W = 10.0, $p = 0.1566$; W = 12.0, $p = 0.0604$).

\end{document}